\pdfoutput=1 

\documentclass[letterpaper]{article} 
\usepackage[preprint]{aaai2027} 
\usepackage[hyphens]{url} 
\usepackage{graphicx} 
\usepackage{natbib} 
\usepackage{caption} 
\usepackage{booktabs}
\usepackage{multirow}

\usepackage{amsmath}
\usepackage{amsfonts}
\usepackage{amsthm}

\theoremstyle{definition}
\newtheorem{definition}{Definition}

\newcommand{\ModelName}{\textsc{Narcissus}}

\newcommand{\hide}[1]{}

\newcommand{\todo}[2][]{}
\newcommand{\tiltodo}[1]{}
\newcommand{\sebtodo}[1]{}
\newcommand{\seb}[1]{}
\newcommand{\nys}[1]{}

\title{\ModelName{}: Program Synthesis Using Context-Aware LLM Approximations}

\author{
    Tilman Hinnerichs,
    Sebastijan Duman\v{c}i\'{c},
    Neil Yorke-Smith
}
\affiliations{
    Delft University of Technology, The Netherlands\\
    \{t.r.hinnerichs, s.dumancic, n.yorke-smith\}@tudelft.nl
}

\begin{document}

\maketitle

\begin{abstract}
    Large language models (LLMs) excel at programming, but not when the task fixes the target language: prompted with a grammar rare in their training data, their programs usually break the grammar or fail the given specification.
Enumerative synthesizers search the space of syntactically correct programs systematically guided by LLMs; the state of the art guides them by approximating LLM proposals into rule frequencies, which loses \emph{where} each construct belongs and prunes every rule the proposals miss, exactly when the proposals are wrong.
We present \ModelName{}, a synthesizer that keeps the proposals as syntax trees and scores each expansion of a candidate program in its \emph{context}: does a proposal with the same surrounding structure continue the same way, and does the expansion rebuild a fragment the proposals repeat?
A regularization term keeps every rule reachable, so wrong proposals delay the solution but cannot hide it.
Across five domains and two search backends, \ModelName{} beats static guidance at every budget and consistently outperforms re-prompting the LLM to fix its own proposals; it reaches proposal-like programs an order of magnitude sooner and solves $40\%$ of ARC tasks where the raw proposals solve $13\%$, all without a single LLM call during search.

\end{abstract}

\section{Introduction}

Large language models (LLMs) have proven to excel at a wide range of programming tasks.
The typical workflow is a loop: prompt the model with the task, sample a program from it, run the program, and check the result; if the program is wrong, sample again or feed the failure back~\cite{AlphaCode,SelfDebug,Reflexion}.
For everyday programming this loop usually suffices: common languages dominate the model's training data, so a correct program is only a few prompts away.

\begin{figure*}[tp]
    \centering
    \includegraphics[width=\textwidth]{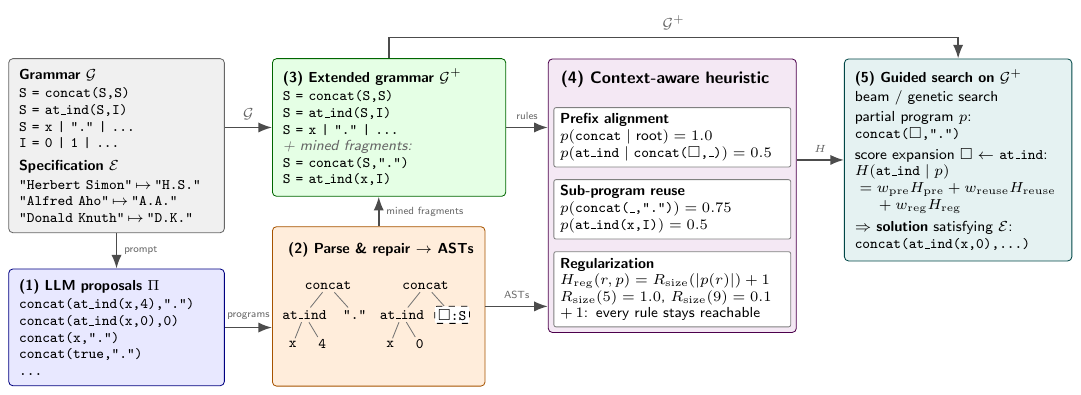}
    \caption{\textbf{Overview of the \ModelName{} pipeline} (on the SLIA task of abbreviating names to initials).
    (1)~The LLM is prompted once with the grammar~$\mathcal{G}$ and specification~$\mathcal{E}$ for a handful of proposals~$\Pi$, which are neither guaranteed solutions nor grammatical (e.g.\ \texttt{concat(true,".")}).
    (2)~Proposals are parsed into abstract syntax trees (ASTs) and repaired against the grammar (syntactically wrong sub-terms become typed holes); (3)~recurring fragments are mined into macro-rules, extending the grammar to~$\mathcal{G}^{+}$; (4)~the proposal ASTs yield a \emph{context-aware} heuristic (prefix alignment along the root-to-hole path, where \texttt{\_} marks a position the context ignores; sub-program reuse; regularization); and (5)~heuristic-based search (e.g., bottom-up or genetic) enumerates programs in~$\mathcal{G}^{+}$, scoring each expansion in its context, and returns the first candidate satisfying~$\mathcal{E}$.
    }
    \label{fig:alg_overview}
\end{figure*}

The loop breaks when the task fixes the language the program must be written in.
Such tasks are everywhere: a compiler for a processor may only emit that processor's instructions, a tool inside a spreadsheet only formulas the spreadsheet understands~\cite{GulwaniFnT}, and abstract-reasoning benchmarks are attacked with hand-designed domain-specific languages~\cite{HodelARCDSL}; a program outside the language cannot run at all.
These specific languages are rare in the LLM's training data, and a grammar in the prompt does not replace training on it: asked for a program in the grammar of Figure~\ref{fig:alg_overview}, even a strong LLM reaches for familiar string functions (\texttt{substring}, \texttt{split}, \texttt{upper}) the grammar does not provide.
Retrying does not fix this: each attempt is a fresh and expensive model call drawn from the same distribution, so the loop mostly resamples the same mistakes.
Programs sampled this way frequently break the grammar or fail the examples~\cite{HySynth,GuidingEnumerativeLLM}; in our benchmarks, even the best proposal model solves at most a third of the tasks outright.
We call the programs an LLM emits for a task its \emph{proposals}.

Program synthesis formalizes this setting: given a grammar defining the language and a specification, e.g., a set of input-output examples, find a program of the grammar that satisfies it.
The LLM of the loop above already acts as a synthesizer, just one that is free to ignore the grammar.
Enumerative synthesizers~\cite{GulwaniFnT,SyGuS2019,EUSolver} instead search the language systematically: they build candidate programs, from now on \emph{candidates}, from the grammar and test each finished one against the examples.
Their weakness is order: the number of programs grows exponentially with program size, so within any budget, success depends on which candidates are tried first.
The na\"ive way to combine them keeps the LLM in the loop, asking at every expansion which rule fits the program built so far; conditioning on context is exactly what makes LLMs strong.
But a guided synthesizer may expand millions of partial candidates per task, so the LLM's judgment must be compiled into something cheaper.

Recent work approximates the proposals with a heuristic.
HySynth~\citep{HySynth} and \citet{GuidingEnumerativeLLM} prompt the LLM a handful of times per task before search and compile the returned proposals into a \emph{static}, context-free prior: each grammar rule scores by how often the proposals use it, independent of the context where it is used.
However, the proposals also strongly indicate \emph{where} a construct belongs; we call the rules already placed around an open position the \emph{context} of the decision that fills it.
For example, if every proposal program ends with \texttt{concat(x,~".")}, then \texttt{concat} is strong evidence at the end of a program and none elsewhere, and the whole sub-term is worth rebuilding together; a plain frequency count captures neither.
Even worse, a rule no proposal uses receives negligible weight, locking the search out of it, exactly when the proposals are wrong and the search would have to correct them.
The compiled guidance is thus blind to where constructs belong and blocked from what the proposals missed. 
Ideally, a heuristic built from proposals must be aware of the context, cheap, and still functional when the proposals are poor.

%
%
We present \ModelName{}, a synthesizer using context-aware LLM approximations as guidance. 
\footnote{Like its namesake, \ModelName{} is drawn to a reflection: it follows the proposals the LLM leaves behind, not the LLM itself.}
During search, \ModelName{} ranks every rule by how often the proposals use it \emph{in the same context}.
To do so, \ModelName{} samples proposals once per task and parses them into abstract syntax trees (ASTs), repairs the parts that violate the grammar, and mines their repeated fragments into shortcut rules (Figure~\ref{fig:alg_overview}).
Three signals then score each expansion of a candidate: \emph{prefix alignment}, whether proposals that share the candidate's context use the same rule; \emph{sub-program reuse}, whether the rule rebuilds a fragment the proposals repeat; and \emph{regularization}, a soft bias towards the proposals' program size plus a positive floor that keeps every rule reachable.
This guidance is cheap: it costs a tree lookup per expansion, and the LLM is never called during search.
Further, since no rule is ever pruned, tasks remain solvable even when the proposals are misleading.

We evaluate \ModelName{} with two backends, a cost-based bottom-up beam and a genetic top-down search, across five domains (SLIA, BV, DeepCoder, ARC, ARGA)~\cite{SyGuS2019,SyGuSLang,DeepCoder,ARCChollet,ARGA}, with proposals from several LLMs ranging from mostly correct to almost never grammar-valid.
The key finding is that context drives the gain: \ModelName{} beats the static prior on all benchmarks under both backends, and plain re-prompting, the loop we opened with, by a wide margin, staying reliable even when few proposals are grammar-valid or correct.
Our heuristic reaches proposal-like programs about an order of magnitude sooner, and on ARC it solves $40\%$ of the tasks while the raw proposals, kept to the target grammar, solve $13\%$.

In summary, we contribute
(i)~a context-aware heuristic that scores each grammar choice by prefix alignment, sub-program reuse, and regularization, cheaply enough to consult at millions of expansions;
(ii)~a pipeline that turns raw LLM text into grammar-aware guidance, repairing ill-formed proposals and mining their recurring fragments into grammar extensions; and
(iii)~evidence across five domains and two backends that context-awareness improves over static LLM-guided synthesis and over prompting the LLM directly, and that regularization makes poor proposals safe.

\section{Preliminaries}

\paragraph{Program Synthesis.}
Program synthesis is the task of finding a program that meets a given specification.
The programs come from a target language $\mathcal{L}(\mathcal{G})$, defined by a context-free grammar $\mathcal{G}$ whose rules specify how the language's operators, constants, and variables combine.
We study inductive program synthesis, the most common one in enumerative synthesis~\cite{GulwaniFnT}, where the specification is a set of input--output examples
$\mathcal{E} = \{(i_1,o_1),\ldots,(i_k,o_k)\}$
and a program $p \in \mathcal{L}(\mathcal{G})$ solves the task if $p(i_j)=o_j$ for all $j$.
Checking a given candidate against $\mathcal{E}$ is cheap; finding one that passes is not.

Enumerative search iteratively builds programs from the grammar, tests each complete one against the specification, and stops when one passes.
Enumeration can proceed top-down, repeatedly expanding open positions starting from the grammar's start symbol so that every search state is a partial program, or bottom-up (BUS), repeatedly composing complete sub-terms into larger ones.
We use variants of both search approaches.
Since the number of programs grows exponentially with program size, most synthesizers rely on a heuristic that decides which candidates to explore first.

\section{Problem Statement}

A search state is a program that may still be partial, $p(p_1,\ldots,\square,\ldots,p_m)$, with open positions called \emph{holes}; it is complete once no hole remains.
Only complete programs can be tested against $\mathcal{E}$.
Choosing a grammar rule $r$ to fill a hole $\square$ is an \emph{expansion}, and a heuristic decides which expansions the search explores first.

The problem we address is to construct this heuristic from LLM proposals: given a grammar $\mathcal{G}$, examples $\mathcal{E}$, and a set of proposals $\Pi$ sampled once before search, build a heuristic that
(i)~scores an expansion depending on the partial program it extends,
(ii)~is cheap enough to consult at every one of millions of expansions without querying the LLM, and
(iii)~leaves every rule of $\mathcal{G}$ reachable, so the guided search still has a chance to solve the task even when $\Pi$ is misleading.
In line with prior work~\citep{HySynth,GuidingEnumerativeLLM}, the proposals serve as guidance rather than answers.
Where we differ is requirement~(i): a static grammar prior assigns each rule one global score, so a rule scores the same wherever it is used.
We instead let the score depend on the partial program the rule expands:

\begin{definition}[Context-aware heuristic]
Given a grammar $\mathcal{G}$ and a set of LLM proposals $\Pi = \{\pi_1,\ldots,\pi_n\}$ for the task, a context-aware heuristic is a function $H(r \mid p, \Pi)$ that scores filling the open position of a partial program $p$ with a rule $r$ using the proposals $\Pi$, where the score of $r$ depends on $p$ \emph{and} $r$.
\end{definition}

Any grammar-based search over $\mathcal{L}(\mathcal{G})$ can use such a heuristic to order its exploration; requirement~(iii) constrains that freedom: the heuristic may reorder the search arbitrarily, but it must leave every part of the grammar reachable, so that misleading proposals delay the solution rather than hiding it entirely.

\section{\ModelName{}}
\label{sec:narcissus}

\ModelName{} turns LLM proposals into a \emph{context-aware heuristic} that any guided search can follow.
Its input is a synthesis task, consisting of a grammar $\mathcal{G}$ and a specification given as input-output examples $\mathcal{E}$, together with access to an LLM.
Before search, \ModelName{} samples a small set of proposal programs $\Pi$ from the LLM and compiles them into the heuristic $H(r \mid p, \Pi)$; the search then uses that heuristic to find a program in $\mathcal{G}$ that satisfies $\mathcal{E}$, without re-querying the LLM.

Throughout, we call the quality of the proposals for a given task the task's \emph{proposal support}: how much useful structure the proposals carry for that specific problem and grammar, quantified in the experimental setup as the fraction of tasks the raw proposals already solve.
\ModelName{} is designed to exploit high proposal support and to degrade gracefully when support is low, never doing worse than unguided search.

\paragraph{Overview.}
\ModelName{}' pipeline (Figure~\ref{fig:alg_overview}) has five steps:
(1)~sample proposals by prompting the LLM once with the specification and the grammar;
(2)~parse the proposal texts into abstract syntax trees and repair the parts that do not fit the grammar;
(3)~mine recurring sub-programs and add them to the grammar as macro-rules, yielding the extended grammar $\mathcal{G}^{+}$;
(4)~compile the repaired trees into the context-aware heuristic; and
(5)~run heuristic-guided search, which enumerates programs in $\mathcal{G}^{+}$ exploring the expansions the proposals support first.

\subsection{Sampling and Repairing LLM Proposals}

\paragraph{Sampling proposal programs.}
Before search, we sample $n$ proposals for the task: we prompt the LLM $n$ times with the specification and the grammar, asking each time for a program that satisfies the specification using only grammar constructs (full template in the appendix), and collect the text proposals $\Pi = \{\pi_1,\ldots,\pi_n\}$; the budget $n$ is fixed per domain.
This sampling happens once, up front; during search \ModelName{} never calls the LLM again.

\paragraph{Parsing and repairing proposals.}
We parse each proposal into a syntax tree, but the trees rarely fit the grammar: an LLM proposal may use an operator the grammar lacks, call a function with the wrong number of arguments, or place an expression of the wrong type.
We therefore repair each tree against the grammar, recursively and top-down.
Starting at the root, we match each node against the grammar rules that can produce the type expected at its position, recursing into the children each matching rule prescribes.
If no rule matches, we replace the sub-term with a hole of its expected type $T$ and stop descending (e.g., an operator the grammar lacks becomes a typed hole while the surrounding structure survives).
Only proposals that yield no expression tree at all are discarded.
The heuristic treats the inserted holes as absent evidence: alignment and reuse still match the intact structure around a hole, while the hole itself supports no particular rule.

\paragraph{Adding frequent sub-programs to the grammar.}
After parsing and repair, we mine sub-programs that recur across the proposal trees: every subtree of at least two nodes that occurs at least twice in the repaired proposals 
is added to the grammar as a macro-rule with its root type as return type.
Mining gives the search a shortcut: a fragment the proposals keep rebuilding can be placed in a single expansion instead of being rediscovered over many.
Mined fragments may themselves contain holes; these become \emph{partial} macro-rules whose holes remain open as nonterminal arguments, such as \texttt{at\_ind(x,I)} in Figure~\ref{fig:alg_overview}.
The augmentation comprises all mined fragments, complete and partial alike, and yields the extended grammar $\mathcal{G}^{+}$.
The idea is in the spirit of library learning~\cite{DreamCoder}, except that we obtain the fragments for free from the proposals for the task at hand.
The augmentation changes what the search can reach in few expansions independently of the heuristic; the experiments isolate its effect with a dedicated baseline (augmented BFS).

\subsection{A Context-Aware Heuristic}

\paragraph{Goal of the heuristic.}
The heuristic $H(r \mid p, \Pi)$ is the heart of \ModelName{}: it scores expanding the open position $\square$ of a partial program $p$ with a grammar rule $r$. The higher the score, the more promising the expansion.
The score combines three signals: prefix alignment, sub-program reuse, and regularization. Each captures a different way the proposals can inform, or fail to inform, the choice.

\paragraph{Prefix alignment.}
The first signal asks whether the proposals use the same rule in the same context.
The context of a hole is the chain of rules on the path from the root of the partial program down to the hole.
Let $c(p)$ be the number of proposals that carry the same rule as $p$ at every position along this path (the proposals ``in the same situation''), and let $s_{\mathrm{prefix}}(r,p)$ be those among them that continue with rule $r$ at the hole itself.
The signal is the \emph{conditional} share of proposals that align with choosing $r$, among those that share the context:
\begin{equation*}
H_{\mathrm{prefix}}(r,p) =
\begin{cases}
s_{\mathrm{prefix}}(r,p) \,/\, c(p) & \text{if } c(p) > 0,\\
0 & \text{otherwise.}
\end{cases}
\end{equation*}
The larger this share, the higher the score; a rule that is common in the proposals overall but not in this context gets no support (step~(4) of Figure~\ref{fig:alg_overview} illustrates this).

\paragraph{Sub-program reuse.}
The second signal rewards rules that help rebuild sub-programs the proposals use repeatedly, independently of context.
Let $s_{\mathrm{reuse}}(r)$ be the number of proposals that contain the sub-program $r$ (here, any subtree, including a single node); then $H_{\mathrm{reuse}}(r) = s_{\mathrm{reuse}}(r)/|\Pi|$.
This is the context-free part of the heuristic; unlike a global rule frequency, it also acts over mined fragments and is combined with the context-aware term above.
Reuse helps when the proposals share local parts but disagree on the whole: a fragment like \texttt{concat(S,~".")} recurring inside different structure biases the search towards programs containing it, wherever the search is, even when no full proposal is correct.

\paragraph{Regularization.}
The proposals indicate not only which constructs are plausible, but also roughly how large the solution should be.
The third signal uses that: it pulls the search towards the program size the proposals suggest.
Let $S(\Pi)$ be the multiset of sizes of the repaired proposal ASTs, and score a program of size $k$ by a Gaussian mixture with one component per proposal,
\begin{equation*}
R_{\mathrm{size}}(k) = \tfrac{1}{Z} \textstyle\sum_{t \in S(\Pi)} \exp\!\left(-(k-t)^2 / 2\sigma^2\right),
\end{equation*}
where the bandwidth $\sigma$ is a hyperparameter of \ModelName{} and $Z = \max_{j} \sum_{t \in S(\Pi)} \exp(-(j-t)^2/2\sigma^2)$ normalises the mixture to peak at $1$.
Sizes count base-grammar operations, so a program built from a mined macro-rule counts at its expanded size.
Writing $p(r)$ for the program that results from expanding $p$ with $r$, the signal is
\begin{equation*}
H_{\mathrm{reg}}(r,p) = R_{\mathrm{size}}(|p(r)|) + C .
\end{equation*}
This term concentrates enumeration at the complexity the proposals indicate (cf.\ RQ3) and keeps the search from drifting into ever-larger programs.
The floor $C > 0$ keeps every rule reachable: prefix alignment and reuse can both be zero and the mixture arbitrarily small, so $C$ guarantees a strictly positive score, excluding no rule.
When the proposals carry no signal, every expansion falls back to $C$, scores alike, and the search degenerates to unguided enumeration, \ModelName{}'s worst case.

\paragraph{Combined score.}
The heuristic adds the three signals with fixed weights; since $\Pi$ is fixed per task, we abbreviate $H(r \mid p, \Pi)$ as $H(r,p)$:
\begin{align*}
H(r,p) ={}& w_{\mathrm{prefix}}\, H_{\mathrm{prefix}}(r,p)
+ w_{\mathrm{reuse}}\, H_{\mathrm{reuse}}(r)\\
&+ w_{\mathrm{reg}}\, H_{\mathrm{reg}}(r,p).
\end{align*}
The three weights, the bandwidth $\sigma$, and the floor $C$ are hyperparameters ($w_{\mathrm{reg}}$ also scales $R_{\mathrm{size}}$); a weight sweep on SLIA found equal weights to work best.
We fix the weights and $C$ at $1$ in every experiment (Appendix~\ref{app:hyperparams}).

\subsection{Search: Using the Heuristic During Synthesis}

We evaluate two search paradigms, top-down and bottom-up, with \ModelName{}' heuristic.

\paragraph{Genetic search.}
The first search method is a genetic algorithm, a top-down search: it manipulates whole programs from the root down, so every candidate carries the context the heuristic needs.
We use a stochastic rather than deterministic top-down search because its mutation operator already coincides with a heuristic-scored expansion, so the heuristic drops in unchanged; a deterministic search would serve equally.
A neighbour is produced by resampling one subtree: the mutation site is a typed hole, a rule is drawn for it from the heuristic-weighted distribution, and the rest is regrown; mutating a subtree is thus still scoring a rule in context, and \ModelName{}'s heuristic applies natively.
Mutation also lets the search move through the proposal region freely, keeping useful fragments while changing the structure around them, which helps when the proposals contain correct sub-programs assembled wrongly.

\paragraph{Cost-guided beam search.}
The second search method follows HySynth's cost-based bottom-up enumeration, the dominant strategy in enumerative synthesis, which allows a more direct comparison of heuristics.
However, we replace the unbounded priority queue with a beam: search combines complete sub-programs into larger ones in order of cost, keeping only a fixed number of best-scoring candidates, which caps memory and lets a limited budget reach deeper programs.
Building bottom-up constructs the root last, so a sub-program has no root-to-hole path to condition on.
We therefore anchor both signals on the sub-program itself: prefix alignment becomes the share of proposals containing that sub-program which give it the parent rule under consideration, and reuse accumulates over the sub-programs already combined into the candidate.
Candidates outside the beam are pruned, so the search moves quickly through large grammars but can lose the solution if the heuristic misleads it early; the regularization term, unavailable to a static prior, keeps the beam from tying itself too tightly to the proposals.

\section{Experimental Evaluation}
\label{sec:experiments}

We evaluate not only \emph{whether} context-aware guidance solves more tasks than static LLM-guided synthesis, but \emph{why}, structured along four research questions.
\textbf{RQ1:} Does \ModelName{} solve more tasks than static LLM-guided synthesis, across the whole range of proposal support?
\textbf{RQ2:} How do prefix alignment, sub-program reuse, and regularization each contribute?
\textbf{RQ3:} Does \ModelName{} reach and enumerate the proposal subspace more efficiently?
\textbf{RQ4:} How sensitive is \ModelName{} to the choice and quality of the proposal LLM?

\subsection{Experimental Setup}

\paragraph{Domains.}
We evaluate across five domains: SLIA and BV are the string- and bit-vector-manipulation tracks of the SyGuS Challenge~\citep{SyGuS2019,SyGuSLang}; DC is the list-manipulation domain of DeepCoder~\cite{DeepCoder,NeoFengMBD18}; and ARC~\cite{ARCChollet}, over the universal Hodel DSL~\cite{HodelARCDSL}, together with its object-centric subset ARGA~\cite{ARGA} tests abstract visual reasoning.
The domains differ sharply in \emph{proposal support}, the fraction of tasks that at least one of the sampled proposals already solves (Table~\ref{tab:proposals}, appendix): support is \emph{strong} on SLIA ($20$--$31\%$) and on ARGA with GPT-4o ($11\%$), and \emph{weak} on BV ($1\%$), DC ($7\%$), ARC ($0$--$7\%$), and ARGA with DeepSeek ($4\%$).
For BV, ARGA, and ARC we fuse \ModelName{} with EUSolver~\cite{EUSolver}, a divide-and-conquer decomposition, as HySynth does on ARGA; \ModelName{}' heuristic drops into this standard synthesizer architecture unchanged.

\paragraph{Baselines.}
We compare against five baselines: unguided breadth-first enumeration (BFS); an \emph{augmented} BFS over the extended grammar $\mathcal{G}^{+}$ (proposal-mined fragments, no heuristic), isolating the value of the fragments; direct LLM sampling (the raw proposals); \emph{re-prompting}, exactly the sample-and-check loop from the introduction: every failing proposal goes back to the LLM with its parse error or wrong outputs; and static LLM-guided synthesis in the style of HySynth~\cite{HySynth}.
To separate heuristic from backend, we run both beam and genetic search under each heuristic: a static rule-frequency prior (\emph{static}) and our context-aware one (\emph{\ModelName{}}); the static variants isolate the value of context.
All static baselines are fit on the same repaired proposals as \ModelName{} (cf.\ the \ModelName{} section), so differences are attributable to the heuristic alone; the pCFG prior of \citet{GuidingEnumerativeLLM} scores rules by proposal frequency like HySynth's, so the static variants already represent that family.

\paragraph{Metrics and protocol.}
We report tasks solved vs.\ programs enumerated and vs.\ wall-clock time, and summarize each curve by its area (AUC), computed in $\log_{10}x$ space and normalized to the curve's average height as a percentage of the task count ($100\%$ = every task solved instantly; a steeper rise scores higher).
Results of the stochastic genetic methods are means over five seeds.
All methods share the same grammar, proposal budget, and a per-task budget of $10^{6}$ programs or $300$ seconds; experiments are implemented in Julia on top of the Herb.jl program synthesis library~\citep{HinnerichsHerbjl}.
Our implementation, the cached LLM proposals, and all result data are available at \url{https://github.com/Herb-AI/Narcissus/}.
\ModelName{}'s weights and floor $C$ are fixed at $1$ throughout, the uniform setting selected by a weight sweep on SLIA (Appendix~\ref{app:hyperparams}).
Proposals come from DeepSeek-V4-Flash and its reasoning variant, Haiku-4.5 for DC, and from the ${\sim}100$ per-task GPT-4o proposals released by HySynth for the subsets of SLIA ($70$ tasks) and ARGA they cover; for DeepSeek we sample $25$ proposals per task on SLIA and $5$ on ARGA.
Genetic search uses a population of $50$ (best of a sweep over $\{10, 50, 100, 1000\}$; Appendix~\ref{app:hyperparams}); our beam search a pool of $1000$ in place of HySynth's unbounded queue.

\subsection{Results}

\begin{figure*}[t]
    \centering
    \begin{minipage}[t]{0.49\textwidth}
        \centering
        \includegraphics[width=\linewidth]{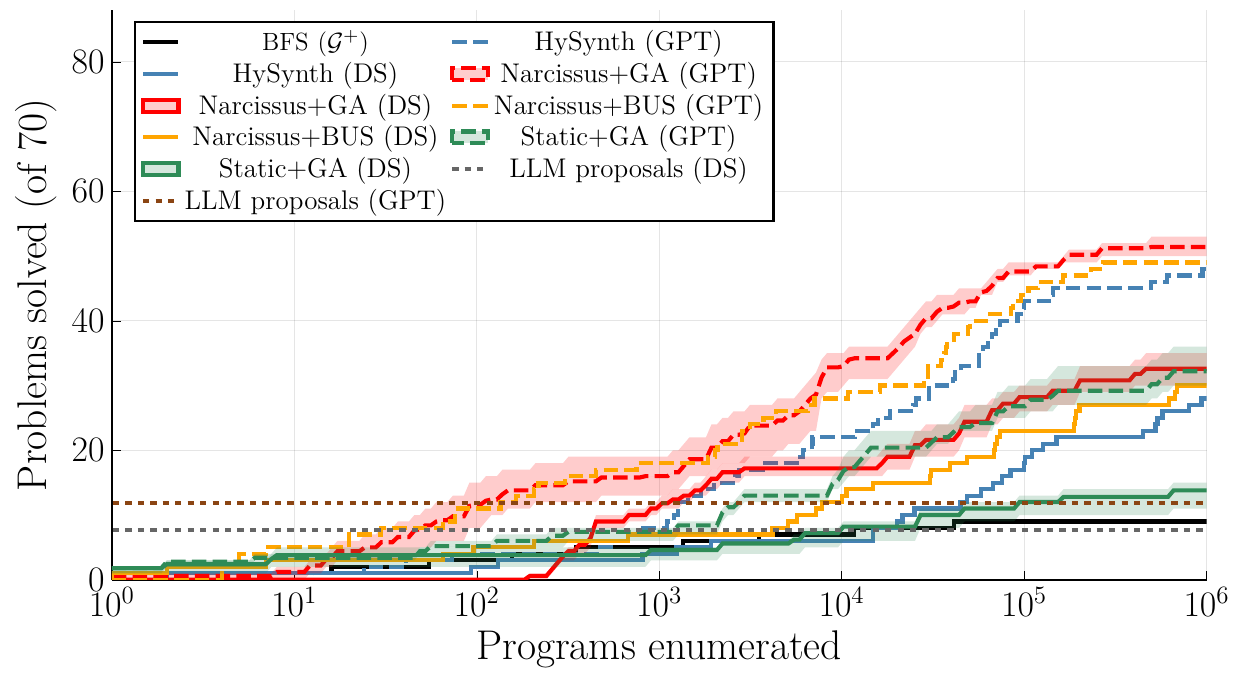}\\
        {\small (a)~SLIA ($70$-task HySynth subset)}
    \end{minipage}\hfill
    \begin{minipage}[t]{0.49\textwidth}
        \centering
        \includegraphics[width=\linewidth]{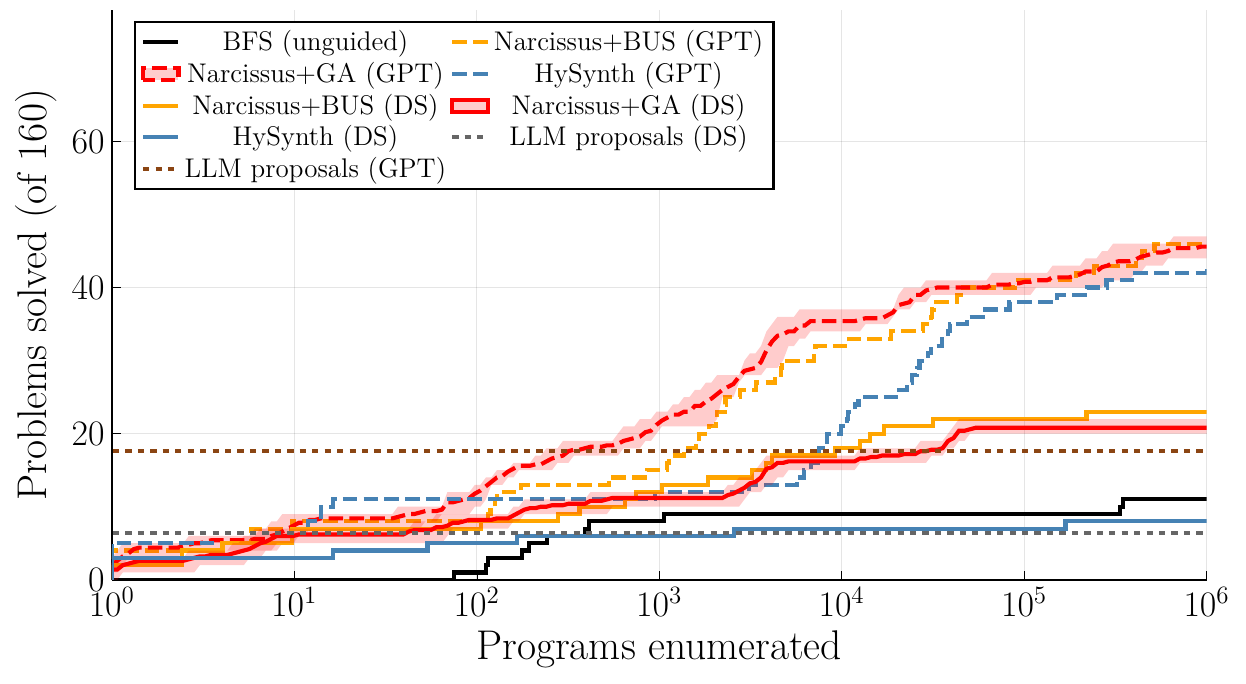}\\
        {\small (b)~ARGA ($160$ tasks, divide-and-conquer)}
    \end{minipage}
    \caption{\textbf{Tasks solved vs.\ programs enumerated, for GPT-4o (GPT, dashed) and DeepSeek-V4-Flash (DS, solid) proposals.} In~(a), \ModelName{} is above the static prior at every budget, under both backends and both proposal models, and its DS curve rivals the static prior on GPT-4o proposals. In~(b), the same ordering holds on a second domain: \ModelName{} beats the static heuristic for both proposal models, well above the raw proposals (dotted).}
    \label{fig:rq1}
\end{figure*}

\begin{table}[tbp]
    \footnotesize
    \setlength{\tabcolsep}{4pt}
    \centering
    \begin{tabular}{lcc}
        \toprule
        Method (proposal model) & Solved (/70) & AUC \\
        \midrule
        \ModelName{} + Genetic (GPT-4o)     & \textbf{51.4} & \textbf{32.2\%} \\
        \ModelName{} + BUS (GPT-4o)         & 49.0 & 30.7\% \\
        Static + BUS (GPT-4o)               & 48.0 & 23.5\% \\
        Static + Genetic (GPT-4o)           & 32.2 & 18.2\% \\
        \midrule
        \ModelName{} + Genetic (DeepSeek)   & \textbf{32.6} & \textbf{17.4\%} \\
        \ModelName{} + BUS (DeepSeek)       & 30.0 & 14.9\% \\
        Static + BUS (DeepSeek)             & 28.0 & 10.5\% \\
        Static + Genetic (DeepSeek)         & 13.8 & \phantom{0}9.1\% \\
        \bottomrule
    \end{tabular}
    \caption{\textbf{SLIA (70 tasks): GPT-4o vs.\ DeepSeek proposals, by final tasks solved and by normalized AUC over programs enumerated.} Higher AUC is better; genetic rows are averaged over 5 seeds, and the best value per proposal model is in bold. At each proposal quality \ModelName{} has the highest AUC, its AUC lead over the static heuristic exceeds its final-count lead (it reaches solutions sooner), and \ModelName{} on DeepSeek proposals matches the static prior on GPT-4o proposals.}
    \label{tab:rq5-auc}
\end{table}

\subsubsection{RQ1: Context-awareness beats static guidance at every level of proposal support.}

We move through the domains from strongest to weakest proposal support.
\emph{Strong proposals (SLIA).}
Figure~\ref{fig:rq1}(a) uses the $70$ SLIA tasks for which HySynth released its GPT-4o proposals, reused unchanged, so we test against HySynth's own guidance signal; our static-prior runs roughly reproduce the numbers the original paper reports.
At every enumeration budget, under either backend and either proposal model, the context-aware heuristic outperforms the static prior, enumerating about a tenth of programs and in roughly a tenth of the time (wall-clock plots in Figure~\ref{fig:rq1-slia-llm}, appendix).
This holds even when we ``starve'' the proposals: our DeepSeek runs use only $25$ per task against HySynth's ${\sim}100$, each a third the size ($14$ vs.\ $44$ rules), yet where the static prior suffers visibly, \ModelName{} barely does.
Re-prompting is no substitute for search either: three rounds of feeding each failing proposal back to the LLM lift the raw DeepSeek proposals by only three tasks.
The full $100$-task set (Figure~\ref{fig:rq1-slia100}, appendix) attributes the gain to context, not the fragments: fragments alone (augmented BFS, $28$) and a static prior ($28.4$) solve about equally many, while \ModelName{} reaches $41.8$ (cf.\ RQ3, RQ4).

\emph{Weak proposals (BV, DC).}
At the other end of the range, guidance must at least do no harm: when the proposals carry little usable signal, guided search should not fall behind no guidance at all.
On BV (Figure~\ref{fig:rq1-bv}, appendix) \ModelName{} solves $350$ of $587$ tasks, above unguided enumeration at $302$, while the static priors collapse to $102$ (BUS) and $57$ (genetic).
The collapse comes from pruning: the static prior makes every rule the weak proposals miss too expensive to enumerate, turning a guide into a constraint; HySynth avoids this only by sampling ${\sim}100$ proposals per task, enough to touch most of the grammar.
\ModelName{}'s regularization instead keeps every rule alive, so its worst case is unguided enumeration; DC tells the same story under both of its proposal models (genetic-\ModelName{} at $32.6$ and $30.6$, BFS at $10$).

\emph{Hard tasks, large grammar (ARC, ARGA).}
The abstract-reasoning benchmarks combine both difficulties: the proposals are weak \emph{and} the tasks are hard, and on ARC the universal Hodel grammar~\cite{HodelARCDSL} spans $320$ rules, so unguided enumeration solves just $4$ of $100$ tasks.
Even here \ModelName{} stays ahead of the static heuristic: on ARGA it wins under both proposal models ($45.6$ vs.\ $43.0$ of $160$ tasks with GPT-4o, $20.8$ vs.\ $8.0$ with DeepSeek), and on full ARC it solves $40$ of $100$ tasks against $10$ and $12$ for the static heuristic under the genetic and BUS backends.
The raw proposals solve $15\%$ of these tasks if any emitted program counts, and $13\%$ once restricted to programs valid in the grammar the synthesizer searches; guided search triples that, because even wrong proposals contribute fragments and structure the search assembles and corrects.
Re-prompting again falls short: on ARGA it solves one more task, reaching $8$ of $160$.

\begin{table}[!tb]
    \footnotesize
    \setlength{\tabcolsep}{4pt}
    \centering
    \begin{tabular}{lcc}
        \toprule
        Signals enabled & DeepSeek (/100) & GPT-4o (/70) \\
        \midrule
        prefix + reuse + reg.\           & \textbf{45.8} & \textbf{51.4} \\
        \midrule
        reuse + reg.\                    & 24.0 & 47.2 \\
        prefix + reg.\                   & 42.8 & 31.4 \\
        prefix + reuse                   & 43.6 & 47.0 \\
        \midrule
        prefix                           & 43.2 & 25.4 \\
        reuse                            & 20.0 & 44.2 \\
        reg.\                            & 13.4 & 15.2 \\
        \bottomrule
    \end{tabular}
    \caption{\textbf{RQ2: ablation of the three heuristic signals, by SLIA tasks solved} (genetic backend, means over 5 seeds; out of $100$ tasks with DeepSeek and out of $70$ with GPT-4o proposals). Rows list the signals left on (\emph{prefix}, \emph{reuse}, \emph{reg.}); the top row is the full heuristic. Which signal is load-bearing inverts with proposal quality: prefix alignment under the weaker DeepSeek proposals, sub-program reuse under the stronger GPT-4o proposals; regularization alone approximates unguided enumeration. Solve curves for all variants are in the appendix (Figure~\ref{fig:rq2-ablation}).}
    \label{tab:rq2-ablation}
\end{table}

\begin{figure}[!t]
    \centering
    \includegraphics[width=\linewidth]{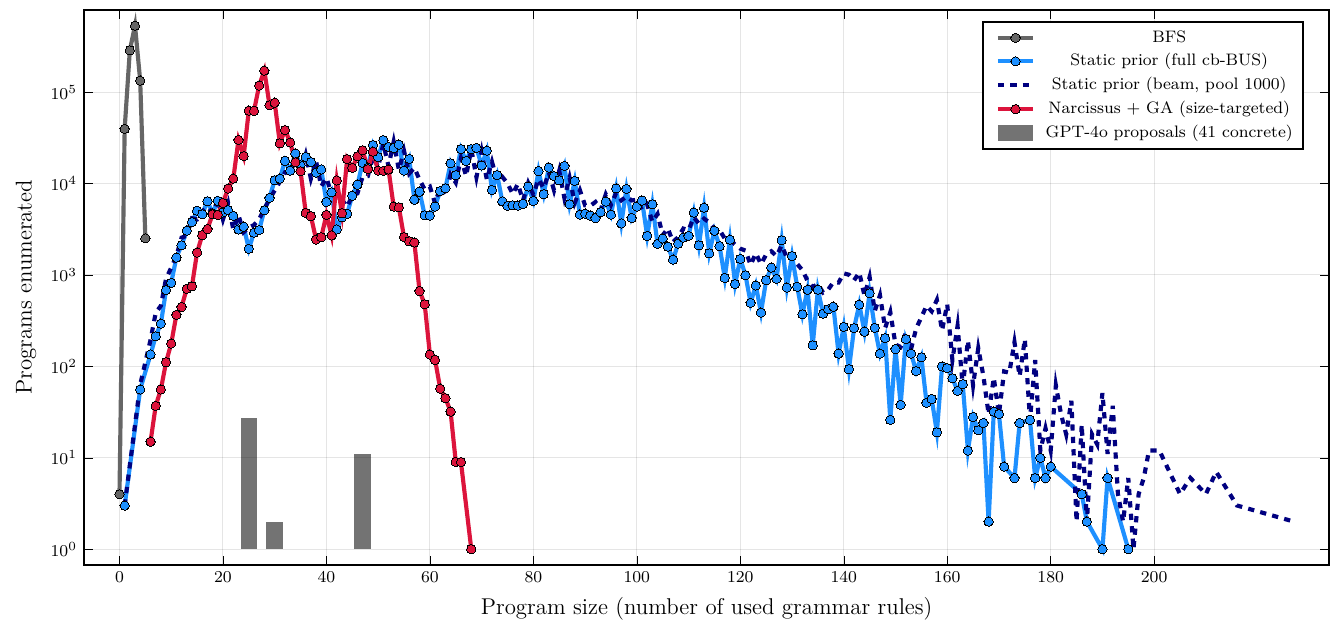}
    \caption{\textbf{Per-size enumeration profile on the SLIA task \texttt{stackoverflow2}} (GPT-4o proposals; grey bars mark the sizes of the $41$ proposals that parse to concrete programs). BFS spends its $10^{6}$-program budget below size five, the static prior (full BUS and width-$1000$ beam, dashed, coinciding) spreads it thinly up to size $195$, and \ModelName{} concentrates it on the sizes the proposals occupy, solving the task alone among the three.}
    \label{fig:rq4-hist}
\end{figure}

\subsubsection{RQ2: Complementary signals; regularization makes weak proposals safe.}

Having traced the gain to the heuristic, we ask which of its three signals carries it, ablating them on SLIA (Table~\ref{tab:rq2-ablation}): all three together, each one left out, and each one alone.
The answer depends on the character of the proposals, and the two proposal models invert each other's story.
DeepSeek proposals are small and largely agree on the structure of the solution, so prefix alignment is decisive: removing it costs the most, dropping the heuristic to $24.0$ solved tasks, and on its own it already recovers almost the full heuristic.
GPT-4o proposals are about three times larger (cf.\ Figure~\ref{fig:rq4-hist}), and at that size the same root-to-hole context rarely recurs across proposals, so prefix alignment fades and sub-program reuse takes over, much as plain rule frequency carries the static heuristic on an un-refined grammar.
BV, from RQ1, supplies the remaining weak-support case: there neither prefix alignment nor reuse has a strong signal to offer, and regularization alone is what keeps the search from collapsing to the static prior's fate.
No single signal dominates everywhere, yet in each setting exactly one is load-bearing; \ModelName{} carries all three at once and so handles every case without knowing in advance which one it will face.

\subsubsection{RQ3: \ModelName{} reaches the proposal subspace faster and at the right complexity.}

To show the mechanism behind RQ1, we measure how many programs a search enumerates before it first reproduces any proposal.
Reaching one is not the goal (seeding would achieve that at step zero); from a cold start it reveals whether the heuristic pulls the search towards proposal-like structure.
It does: on SLIA with DeepSeek proposals, \ModelName{} reaches the proposal region ${\approx}12\times$ sooner (in programs enumerated) than the static heuristic.
Figure~\ref{fig:rq4-hist} makes the mechanism concrete on a single task, plotting how each method distributes its enumeration budget over program sizes: regularization concentrates \ModelName{}'s budget on the sizes the proposals indicate, while BFS piles onto tiny programs and the static prior spreads a long tail well past the proposals, neither landing where the solution lies.

\subsubsection{RQ4: Search recovers much of the gap between cheap and strong proposals.}

Finally, we vary the proposal model (Figure~\ref{fig:rq1}(a), Table~\ref{tab:rq5-auc}).
A cheap model plus search beats an expensive model alone: sampled directly, GPT-4o solves $31\%$ against DeepSeek-V4-Flash's $20\%$, yet \ModelName{} on the DeepSeek proposals reaches $47\%$ ($32.6$ of $70$), overtaking direct GPT-4o sampling and matching the \emph{static} prior on GPT-4o proposals.
\ModelName{} leads at both qualities, by more in AUC than in final count ($+7.2$ vs.\ $+1.0$ over the static heuristic at GPT-4o): it extracts more from a fixed proposal set, and sooner.

\section{Related Work}

\paragraph{Scaling the sample-and-check loop.}
The most direct answer to a failing proposal is more LLM work, in three flavours.
The first simply samples more programs and keeps those that pass the examples, on ARC up to thousands per task~\cite{Codex,AlphaCode,ARCPrize2024,BARC}.
Self-repair feeds execution feedback back to the model~\cite{SelfDebug,Reflexion}; our re-prompting baseline evaluates exactly this and gains a single task on ARGA.
Constrained decoding forces each generated token to follow the grammar~\cite{GrammarAugmentation}, fixing syntax but not correctness, and distorting the model's distribution~\cite{GrammarAlignedDecoding}.
All three keep drawing from a model that does not know the target language; none searches it.

\paragraph{LLM-in-the-loop program search.}
A second group embeds the LLM inside a symbolic search, as generator, mutator, or scorer of candidates~\cite{HypothesisSearch,CodeIt,SOAR,FunSearch,ConceptSearch,GuidingEnumerativeLLM}.
This restores search, but the number of LLM calls now grows with search effort (excluded by our problem statement): these methods spend a model call where \ModelName{} spends a tree lookup.

\paragraph{Static LLM approximations for search.}
Closest to \ModelName{} is work that prompts the LLM once and compiles the samples into fixed search guidance.
HySynth~\cite{HySynth} and \citet{GuidingEnumerativeLLM} compile a context-free heuristic; \citet{TaoGGGP} seed grammar-repaired samples into genetic programming steered by whole-program similarity; Probabilistic Programs of Thought~\cite{PPoT} rebuilds a program distribution from one generation's token probabilities, in the LLM's own language; concurrent ReaComp~\cite{ReaComp} compiles reasoning traces into standalone solvers.
None scores an expansion by its context in a fixed target grammar: their distributions are position-independent and prune what the samples miss; a context-dependent approximation is the future direction \citet{HySynth} themselves name.

\paragraph{Learned search heuristics.}
Guiding enumeration with learned models predates LLMs, from rule predictors and value rankers to library learning~\cite{DeepCoder,BUSTLE,BeeSearch,DreamCoder}; Euphony~\cite{LearnedProbModels} even conditions rule probabilities on surrounding structure, and Probe~\cite{Probe} re-weights the grammar just-in-time during search.
What this group lacks is knowledge of the task at hand: it trains on other tasks, or only on the search's own partial results, whereas \ModelName{} obtains context-aware, task-specific guidance from a handful of proposals, with no training.

\section{Conclusion and Future Work}

We presented \ModelName{}, which turns LLM proposals into a context-aware heuristic: where a static prior asks only \emph{how often} a rule occurs, \ModelName{} asks \emph{where}.
Because no rule is ever pruned, it corrects the LLM rather than re-ranking it, and across five domains and two backends it beats static LLM-guided synthesis with no LLM call during search.
Our weights are fixed at one; a per-task weighting, set from cheap properties of the proposals such as their size and agreement, could lean on whichever of the three signals a task actually supports (cf.\ RQ2), rather than treating all three alike.
Mined fragments are currently discarded after each task, but sharing them across tasks would turn the per-task grammar extensions into a growing, reusable library, and re-seeding the search from its own near-misses could recover solutions that lie just outside the proposal region.
More broadly, the same context-aware approximation applies wherever an LLM can propose but not reliably produce grammatical programs, a cheap alternative to keeping the model in the search loop.

\bibliography{references}

\cleardoublepage

\appendix
\setcounter{secnumdepth}{2}

\section{Intuition: Searching Around the Proposals}
\label{app:intuition}

\begin{figure*}[tp]
    \centering
    \includegraphics[width=\textwidth]{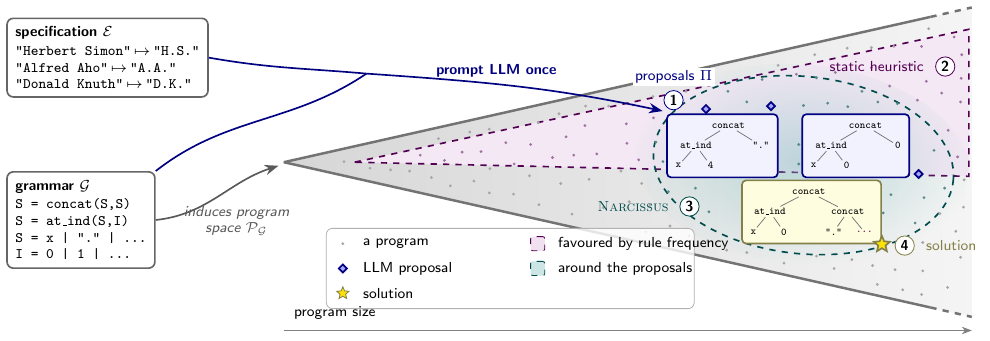}
    \caption{\textbf{Why we search around the proposals}, on the SLIA task of Figure~\ref{fig:alg_overview}.
    The grammar induces the program space $\mathcal{P}_{\mathcal{G}}$, drawn so that program size grows to the right; each dot is one program.
    Prompted once with $\mathcal{G}$ and $\mathcal{E}$, the LLM returns a few raw proposals~(1): wrong, sometimes ungrammatical, but sitting close together in a small region of the space.
    Scoring programs by how often their rules occur in the proposals~(2) covers that region, but with it every program built from the same rules at every size, most of them far larger than anything the proposals suggest.
    What is worth enumerating instead is the region \emph{around} the proposals~(3): their rules, in their places, at their size.
    It has to stay open, because the solution~(4) can need a rule the proposals never used.}
    \label{fig:motivation}
\end{figure*}

Figure~\ref{fig:motivation} illustrates the intuition behind \ModelName{} and how it differs from static LLM-guided synthesis.

\section{Prompt Template}
\label{app:prompt}

Every proposal is sampled with the following prompt; \texttt{<GRAMMAR>} and \texttt{<EXAMPLES>} are replaced per task by the grammar's rules and the input-output examples, respectively.

{\scriptsize
\begin{verbatim}
You are a program synthesizer. Output ONLY one
program in the target DSL that satisfies all
examples. Do not explain. Do not add comments
or extra text. If no solution fits the grammar,
output exactly: (no-solution).

# TASK
Given:
1) A domain-specific language (DSL) as a
   context-free grammar (CFG).
2) A set of input->output examples.

Produce ONE program in the DSL that:
- Is syntactically valid w.r.t. the CFG.
- When executed, returns the required outputs
  for ALL examples.

# OUTPUT FORMAT (STRICT)
Return ONLY the program, nothing else.
- return the complete program only.
- Do NOT include backticks or prose.

# MINIMALITY & VALIDITY
- Prefer the simplest correct program (fewest
  nodes / constructs) when multiple work.
- Use ONLY terminals and productions allowed
  by the provided CFG.
- Do NOT invent identifiers, predicates, or
  library calls outside the CFG.
- Respect arities and types exactly.

# DSL (CFG)
<GRAMMAR>

# SPECIFICATION
# Input->Output examples the program MUST
# satisfy:
<EXAMPLES>

# CONSTRAINTS
- No I/O; compute only via allowed DSL
  primitives.
- Execution must terminate on all provided
  examples.

# SELF-CHECK BEFORE YOU ANSWER (NO OUTPUT OF
# THIS THINKING):
1) Parse your candidate against the CFG.
2) Mentally trace it on ALL examples; confirm
   the outputs match exactly.
3) If any check fails, revise. Otherwise,
   output ONLY the final program.
\end{verbatim}
}

\subsection{Re-prompting Template}
\label{app:reprompt}

The \emph{re-prompting} baseline (\S\ref{sec:experiments}) feeds each failing proposal back to the LLM with what went wrong and asks for one corrected program, for up to three rounds. \texttt{<PREVIOUS>} is the last program the model returned; \texttt{<FEEDBACK>} reports the failure, and \texttt{<ISSUE>} switches framing between an ungrammatical program and a grammatical one with wrong outputs. \texttt{<GRAMMAR>} and \texttt{<EXAMPLES>} are as above.

{\scriptsize
\begin{verbatim}
You are a program synthesizer. You previously
proposed a program in the target DSL, but it
was not correct. Revise it.
Output ONLY one program in the DSL. Do not
explain, no comments, no backticks. If truly no
program fits the grammar, output exactly:
(no-solution).
This is a blind revision: do not execute, run,
or otherwise test the program with any tool --
reason about it purely on paper.

# DSL (CFG)
<GRAMMAR>

# SPECIFICATION
# Input->Output examples the program MUST
# satisfy:
<EXAMPLES>

# YOUR PREVIOUS PROGRAM
<PREVIOUS>

# WHAT WENT WRONG
<ISSUE>
<FEEDBACK>

# INSTRUCTIONS
- Produce ONE corrected program in the DSL that
  satisfies ALL examples.
- Use ONLY terminals and productions allowed by
  the CFG. Respect arities and types exactly.
- Do NOT invent identifiers, predicates, or
  library calls outside the CFG.
- Prefer the simplest correct program.

# OUTPUT FORMAT (STRICT)
Return ONLY the program, nothing else. No
backticks, no prose.
\end{verbatim}
}

\noindent where \texttt{<ISSUE>} is one of:

{\scriptsize
\begin{verbatim}
Your previous program is NOT a valid program in
the DSL grammar above:

Your previous program is a valid DSL program,
but it does not satisfy all the examples. Here
is exactly what it did:
\end{verbatim}
}

\section{Raw Proposal Accuracy}
\label{app:proposals}

Table~\ref{tab:proposals} reports how many tasks the raw proposals already solve, per domain and proposal model.
This is the \emph{proposal support} referred to throughout the main text, and the lower block, which counts only proposals that are syntactically valid in the target grammar, is the accuracy the search results should be compared against.

\begin{table*}[tbp]
    \centering
    \footnotesize
    \setlength{\tabcolsep}{3pt}
    \begin{tabular}{p{1.4cm}p{0.6cm}p{1.5cm}p{1.5cm}p{1.5cm}p{1.5cm}p{1.5cm}p{1.5cm}p{1.5cm}p{1.5cm}p{1.5cm}}
        \toprule
        Group & \(k\) & \textbf{SLIA (Deep\-Seek-Chat)} (N=205) & \textbf{SLIA (GPT-4o)} (N=70) & \textbf{BV (Deep\-Seek-Chat)} (N=753) & \textbf{DC (Deep\-Seek-Chat)} (N=100) & \textbf{DC (Haiku)} (N=100) & \textbf{ARC (Deep\-Seek-Chat)} (N=42) & \textbf{ARC (Deep\-Seek-Reasoner)} (N=100) & \textbf{ARGA (Deep\-Seek-Chat)} (N=160) & \textbf{ARGA (GPT-4o)} (N=160) \\
        \midrule
        \multirow{3}{=}{All proposals} & \(k=1\) & 27 (13\%) & 8 (11\%) & 6 (1\%) & 3 (3\%) & 6 (6\%) & 0 (0\%) & -- & 0 (0\%) & 2 (1\%) \\
         & \(k=5\) & 31 (15\%) & 13 (19\%) & 9 (1\%) & 6 (6\%) & 7 (7\%) & 0 (0\%) & -- & 4 (2\%) & 6 (4\%) \\
         & all & 42 (20\%) & 22 (31\%) & 11 (1\%) & 7 (7\%) & 7 (7\%) & 0 (0\%) & 15 (15\%) & 7 (4\%) & 18 (11\%) \\
        \midrule
        \multirow{3}{=}{Synth. valid only} & \(k=1\) & 18 (9\%) & 5 (7\%) & 6 (1\%) & 3 (3\%) & 3 (3\%) & 0 (0\%) & -- & 0 (0\%) & 2 (1\%) \\
         & \(k=5\) & 19 (9\%) & 9 (13\%) & 9 (1\%) & 3 (3\%) & 3 (3\%) & 0 (0\%) & -- & 4 (2\%) & 6 (4\%) \\
         & all & 22 (11\%) & 12 (17\%) & 11 (1\%) & 4 (4\%) & 3 (3\%) & 0 (0\%) & 13 (13\%) & 7 (4\%) & 18 (11\%) \\
        \bottomrule
    \end{tabular}

    \caption{\textbf{Raw proposal accuracy per domain and proposal model.} We report the number (and share) of tasks for which at least one of the first $k$ proposals satisfies all examples; ``all'' uses every available proposal. The lower block counts only proposals that are syntactically valid in the target grammar, i.e.\ the language the synthesizer must search, which is the accuracy the search results should be compared against. $N$ is the number of tasks for which proposals were collected from that model; it can exceed the evaluation subsets used in the experiments (e.g.\ DeepSeek proposals cover all $205$ SLIA tasks while the search experiments use the standard $100$-task subset, and the GPT-4o column covers the $70$-task subset released with HySynth; the ARC (DeepSeek-Chat) column covers the $42$ tasks prompted for that model). The ARC (DeepSeek-Reasoner) column covers the canonical $100$-task subset used for the full-ARC RQ1 experiment (Figure~\ref{fig:rq3-arc}); only the ``all'' row is available for this subset, since the $k=1$/$k=5$ breakdown has not been recomputed for it.}
    \label{tab:proposals}
\end{table*}

\section{Hyperparameter Selection}
\label{app:hyperparams}

The search has two free hyperparameters: the heuristic's signal weights and the genetic population size. We fix both by small sweeps on SLIA and reuse the chosen values unchanged in every other domain, so the numbers reported throughout the paper use a single, domain-independent configuration.

\paragraph{Signal weights.}
The heuristic combines prefix alignment, sub-program reuse, and regularization with weights $w_{\mathrm{prefix}}, w_{\mathrm{reuse}}, w_{\mathrm{reg}}$ and a floor $C$. Varying each weight over $\{0.5, 1, 5, 10\}$ on SLIA, we found no skewed assignment that consistently improved on giving the three signals equal weight; setting all of them (and $C$) to $1$ solved the most tasks throughout. This matches the RQ2 finding that each signal is load-bearing in a different regime: any assignment that down-weights one signal helps the tasks that rely on the others but hurts those that rely on it, so equal weights are the robust choice. We therefore adopt weights of $1$ everywhere and never re-tune them per domain.

\paragraph{Genetic population size.}
For the genetic backend we swept the population size over $\{10, 50, 100, 1000\}$ on SLIA. Larger populations solve marginally more tasks but cost proportionally more time per generation, exhausting the wall-clock budget on fewer tasks; $50$ gave the best trade-off between final solve rate and running time, and we use it for every genetic run. The beam-search backend analogously uses a pool of $1000$, matching the beam width against which HySynth's unbounded queue is compared.

\section{Additional Results: Solve Curves}
\label{app:curves}

We provide the full set of cumulative solve curves; for the discussion of these results please see the Experimental Evaluation section of the main text.
Each domain is shown on both budget axes: \emph{programs enumerated} (left), which is implementation- and hardware-independent, and \emph{wall-clock time} (right), which additionally charges each method for the cost of its own guidance.
Curves for the stochastic genetic methods are means over five seeds, with shaded bands giving the spread across seeds.

\subsection{RQ1: SLIA}
\begin{figure*}[tbp]
    \centering
    \includegraphics[width=0.49\textwidth]{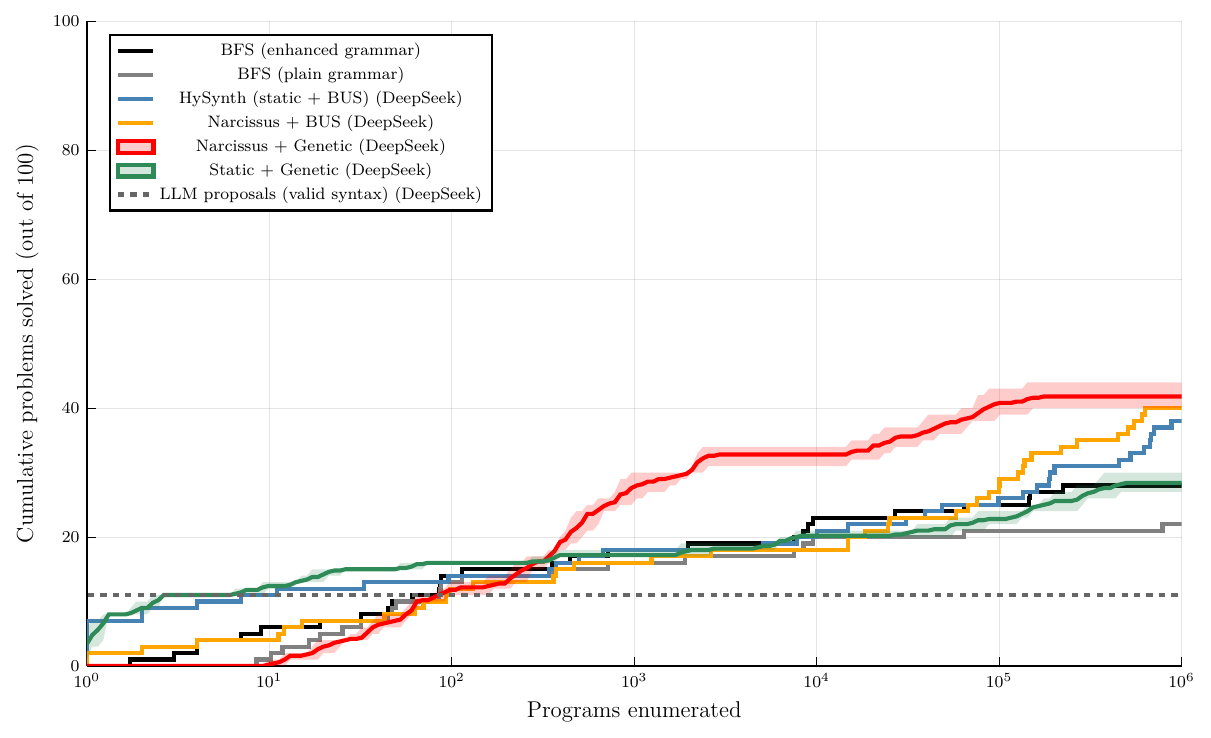}\hfill
    \includegraphics[width=0.49\textwidth]{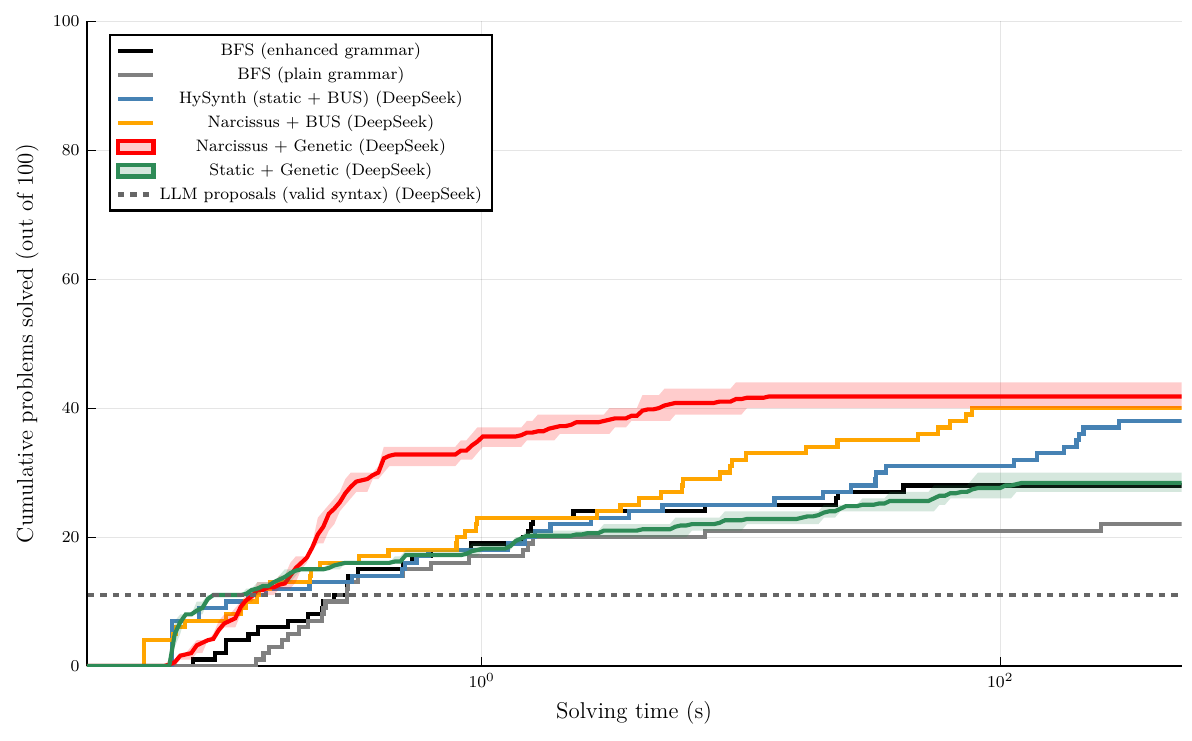}
    \caption{\textbf{RQ1, full 100-task SLIA (DeepSeek proposals): tasks solved vs.\ programs enumerated (left) and vs.\ wall-clock time (right).} With the same backend, \ModelName{} dominates the static prior at every budget on both axes, and reaches solutions orders of magnitude earlier than the static heuristic. The raw grammar-valid proposals (dashed) are shown for reference. The ordering is unchanged when guidance is charged for its own runtime, so the gain is not an artifact of counting enumerations only.}
    \label{fig:rq1-slia100}
\end{figure*}

\begin{figure*}[tbp]
    \centering
    \includegraphics[width=0.49\textwidth]{figures/plots/rq1_slia_llm_comparison_enumerations.pdf}\hfill
    \includegraphics[width=0.49\textwidth]{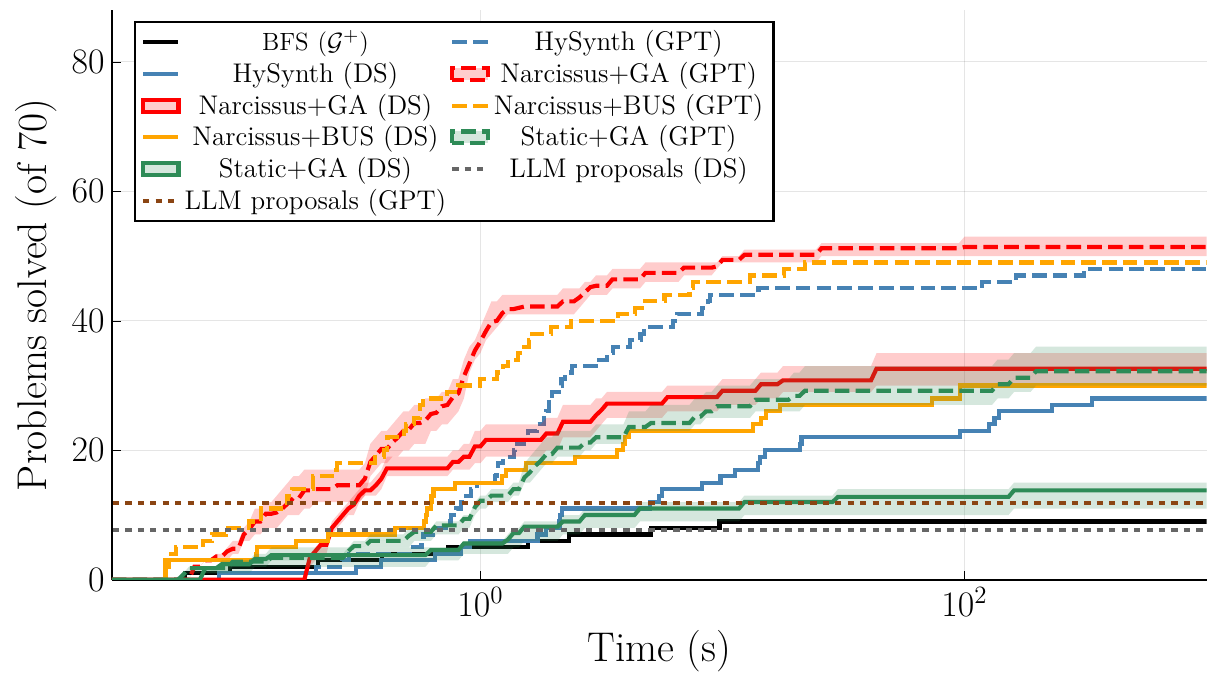}
    \caption{\textbf{RQ1/RQ4, SLIA 70-task subset: GPT-4o vs.\ DeepSeek proposals, by programs enumerated (left) and wall-clock time (right).} This is the time-axis counterpart of Figure~\ref{fig:rq1}(a). At each proposal quality \ModelName{} leads under both backends, and \ModelName{} on the DeepSeek proposals rivals the static prior on the GPT-4o proposals.}
    \label{fig:rq1-slia-llm}
\end{figure*}

Figure~\ref{fig:rq1-slia100} covers the full 100-task set with DeepSeek proposals, and Figure~\ref{fig:rq1-slia-llm} the 70-task HySynth subset under both proposal models; the latter is the wall-clock counterpart of Figure~\ref{fig:rq1}(a).

\subsection{RQ1: Bit-Vectors (BV)}
\begin{figure*}[tbp]
    \centering
    \includegraphics[width=0.49\textwidth]{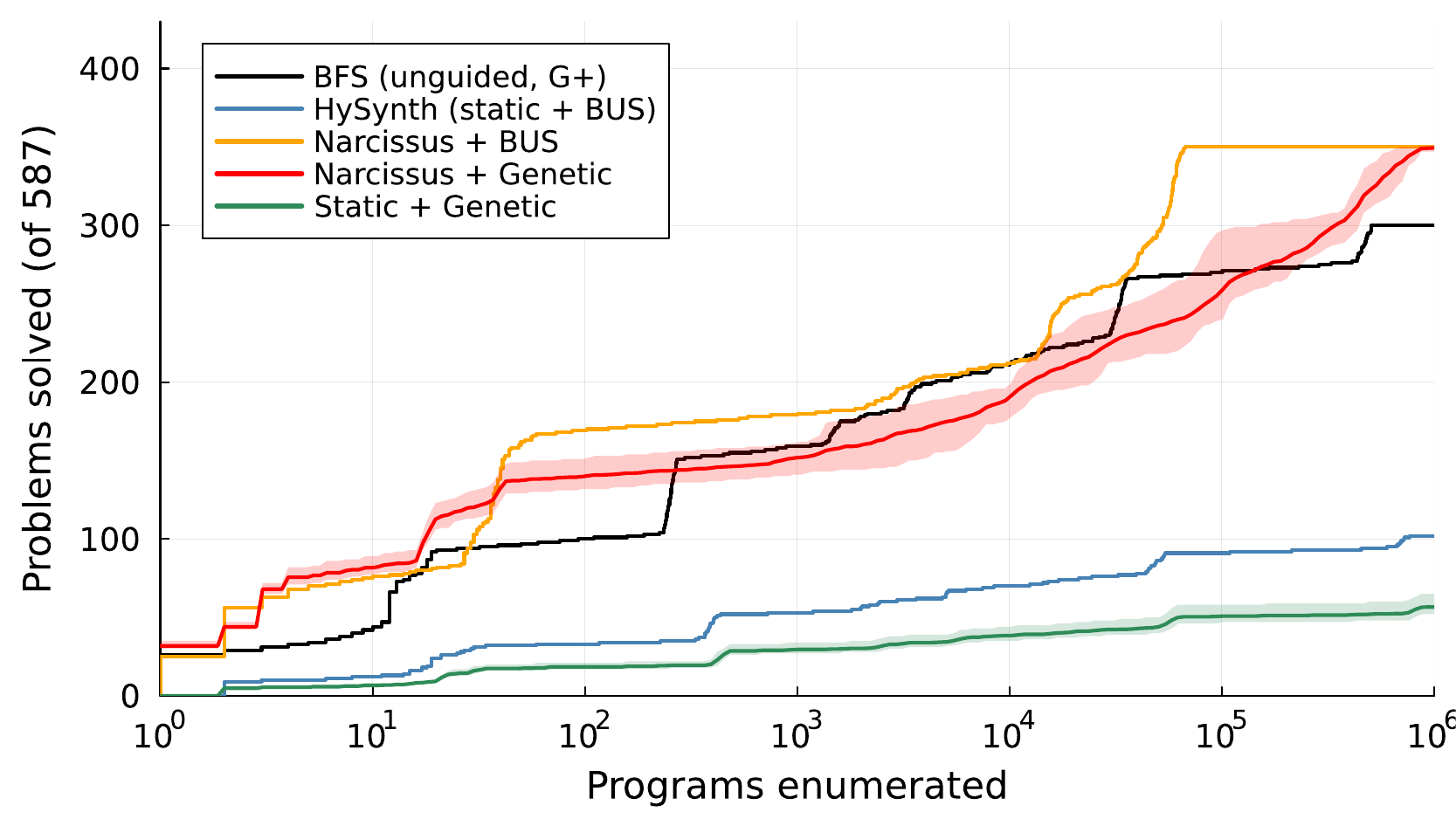}\hfill
    \includegraphics[width=0.49\textwidth]{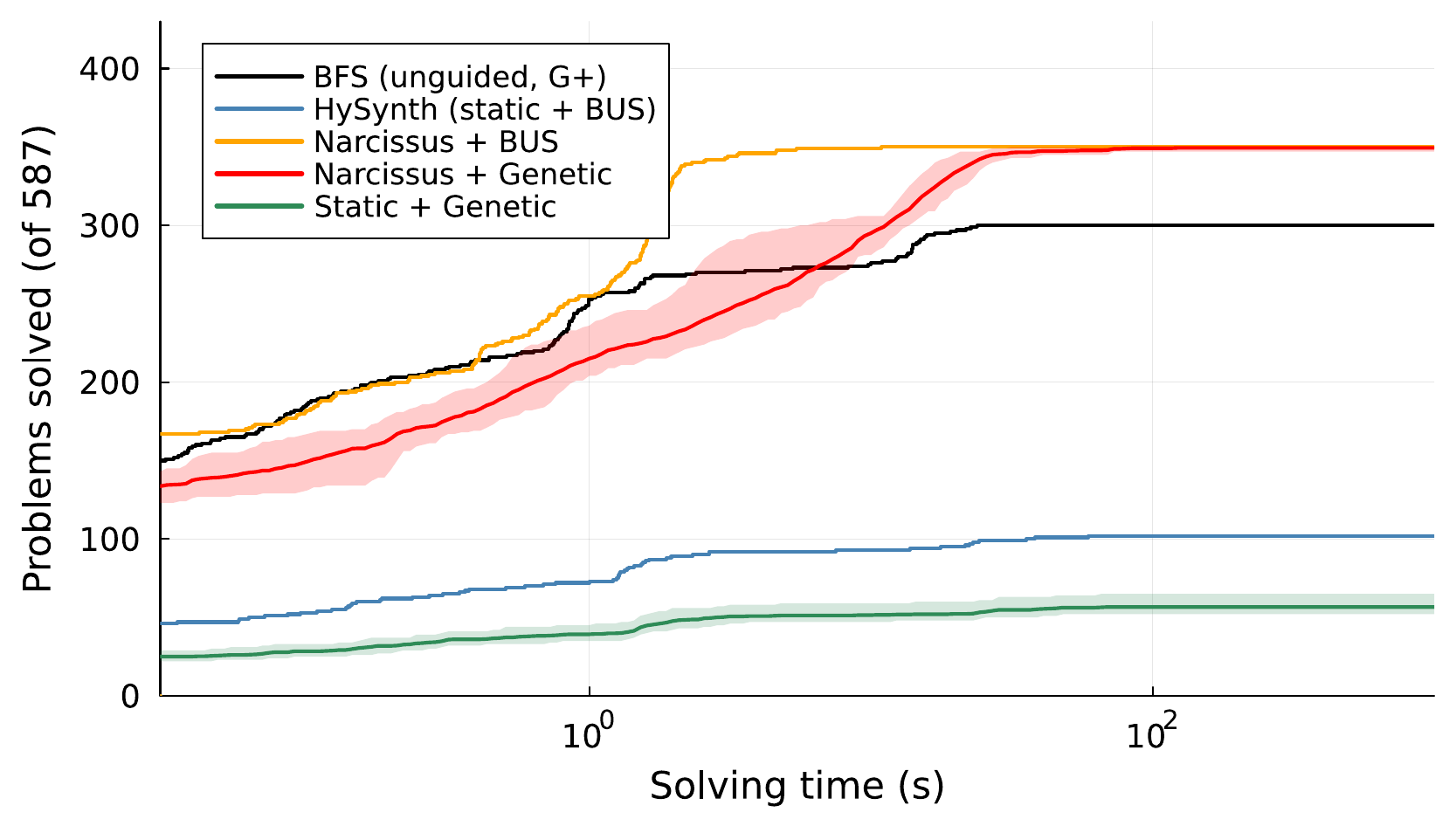}
    \caption{\textbf{RQ1, BV with divide-and-conquer (weak proposals): programs enumerated (left) and wall-clock time (right).} \ModelName{} stays above unguided enumeration, while both static-prior variants collapse far below it, because they effectively prune the rules the weak proposals never mention. This is the core robustness result behind the claim that \ModelName{}'s floor is plain enumeration. The unguided baseline enumerates the extended grammar $\mathcal{G}^{+}$, so it is not handicapped relative to the guided methods.}
    \label{fig:rq1-bv}
\end{figure*}

\begin{figure*}[tbp]
    \centering
    \includegraphics[width=0.49\textwidth]{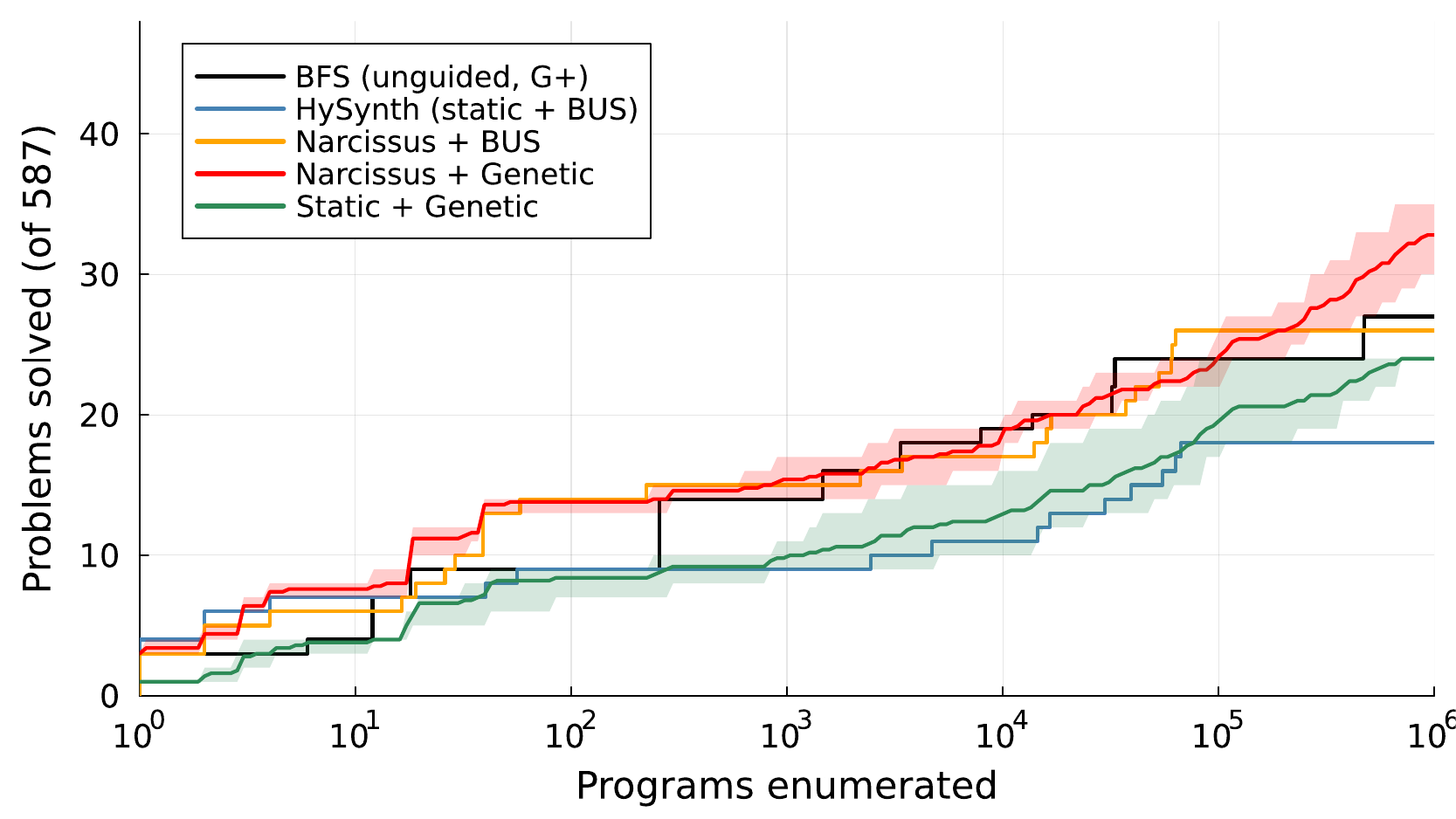}\hfill
    \includegraphics[width=0.49\textwidth]{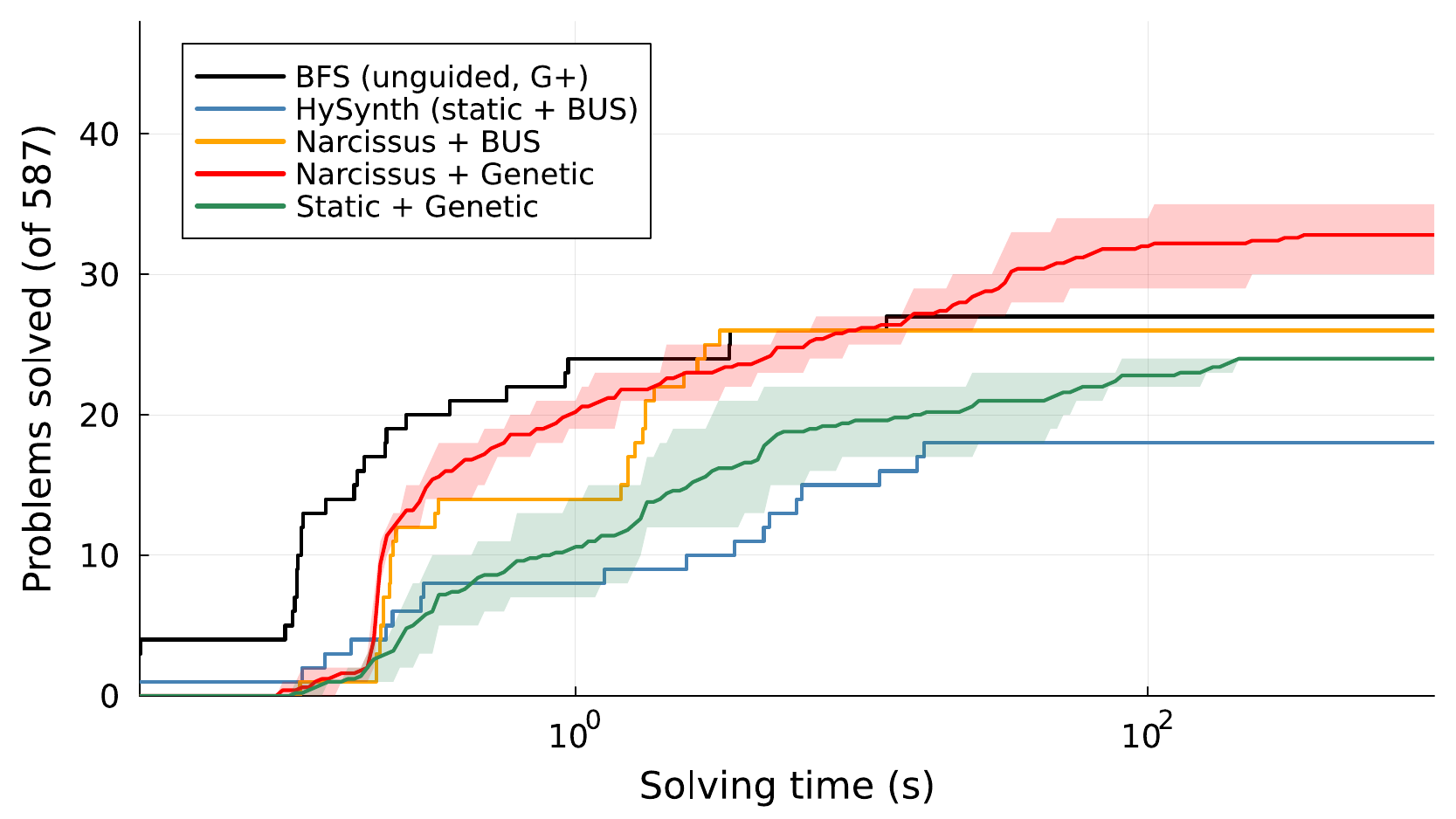}
    \caption{\textbf{RQ1, BV without divide-and-conquer (plain synthesizer): programs enumerated (left) and wall-clock time (right).} Without the divide-and-conquer decomposition every method solves far fewer of the $587$ tasks, so the y-axis is clipped at $48$ to keep the curves legible; the ordering of Figure~\ref{fig:rq1-bv} is largely preserved, with \ModelName{} remaining close to or above unguided enumeration while the static priors fall clearly below it. As in Figure~\ref{fig:rq1-bv}, the unguided baseline enumerates the extended grammar $\mathcal{G}^{+}$.}
    \label{fig:rq1-bv-plain}
\end{figure*}

Figure~\ref{fig:rq1-bv} shows BV under the divide-and-conquer decomposition used in the main text, and Figure~\ref{fig:rq1-bv-plain} the same benchmark without it.

\subsection{RQ1: DeepCoder (DC)}
\begin{figure*}[tbp]
    \centering
    \includegraphics[width=0.49\textwidth]{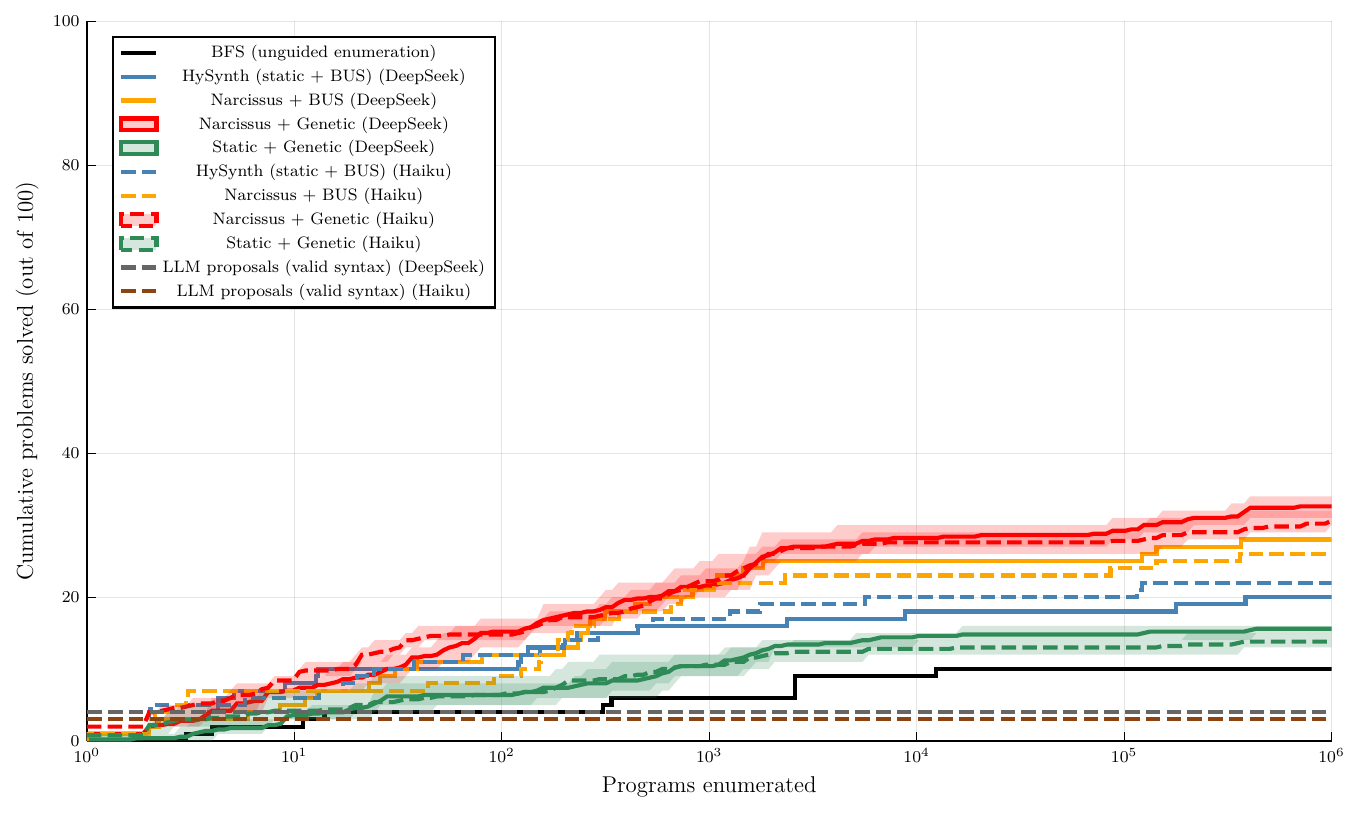}\hfill
    \includegraphics[width=0.49\textwidth]{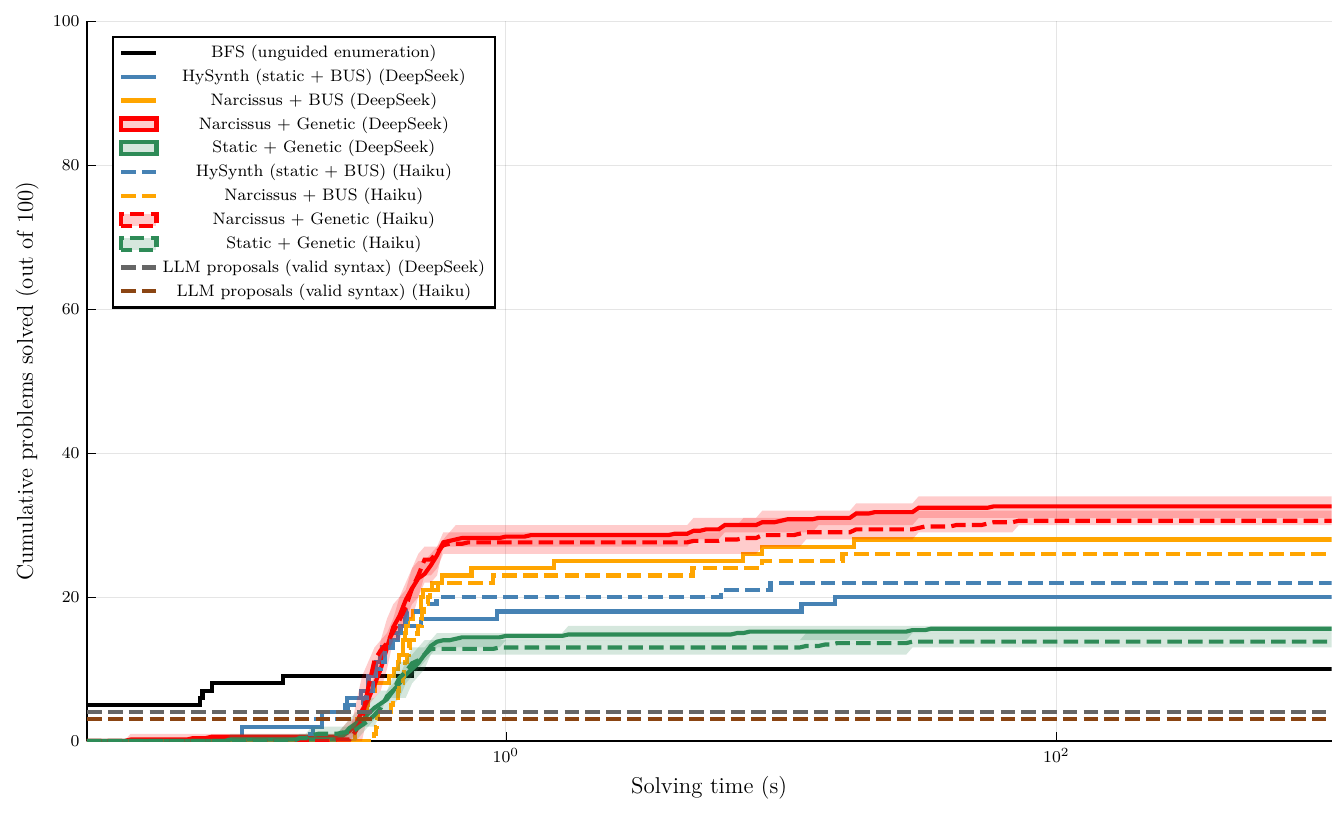}
    \caption{\textbf{RQ1, DeepCoder ($100$ tasks), DeepSeek vs.\ Haiku proposals: programs enumerated (left) and wall-clock time (right).} DC has weak proposal support ($7\%$, Table~\ref{tab:proposals}), yet the RQ1 ordering holds for both proposal models and under both backends: \ModelName{} leads, the static prior trails, and unguided enumeration is the floor.}
    \label{fig:rq1-dc}
\end{figure*}

Figure~\ref{fig:rq1-dc} compares DeepSeek and Haiku proposals on the 100 DeepCoder tasks.

\subsection{RQ2: Signal Ablation}
\begin{figure*}[tbp]
    \centering
    \includegraphics[width=0.49\textwidth]{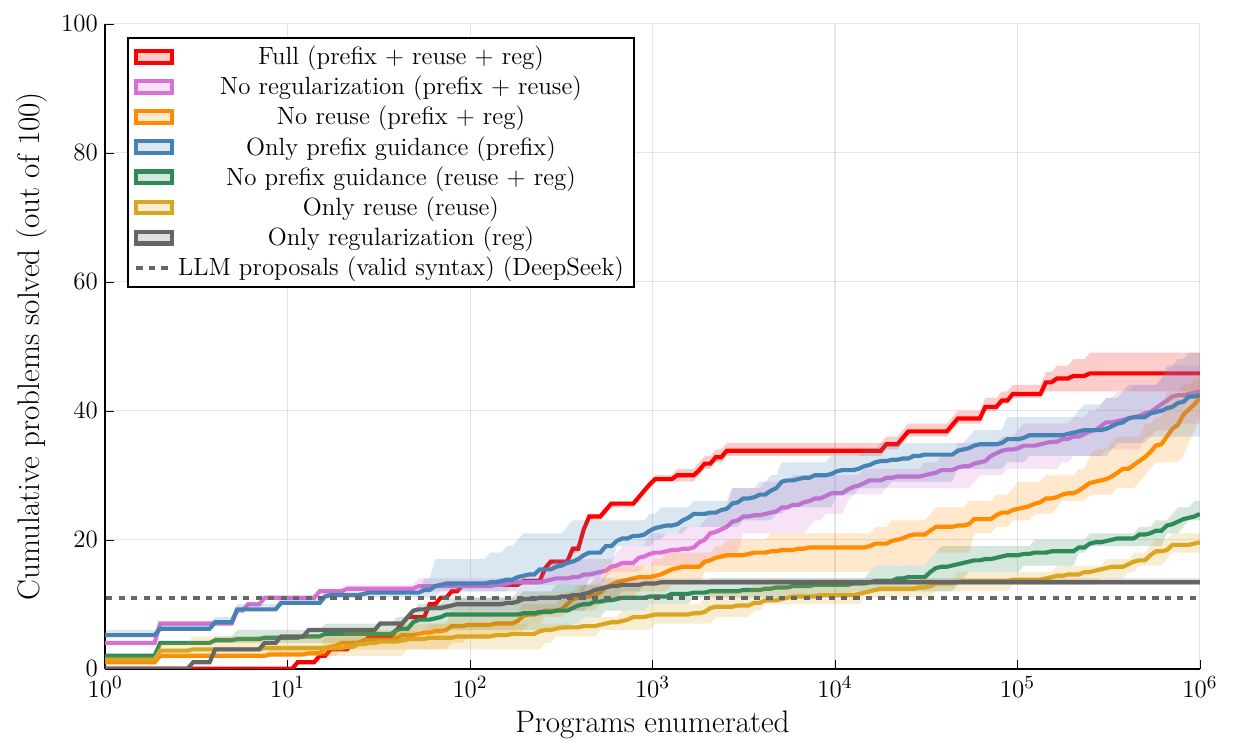}\hfill
    \includegraphics[width=0.49\textwidth]{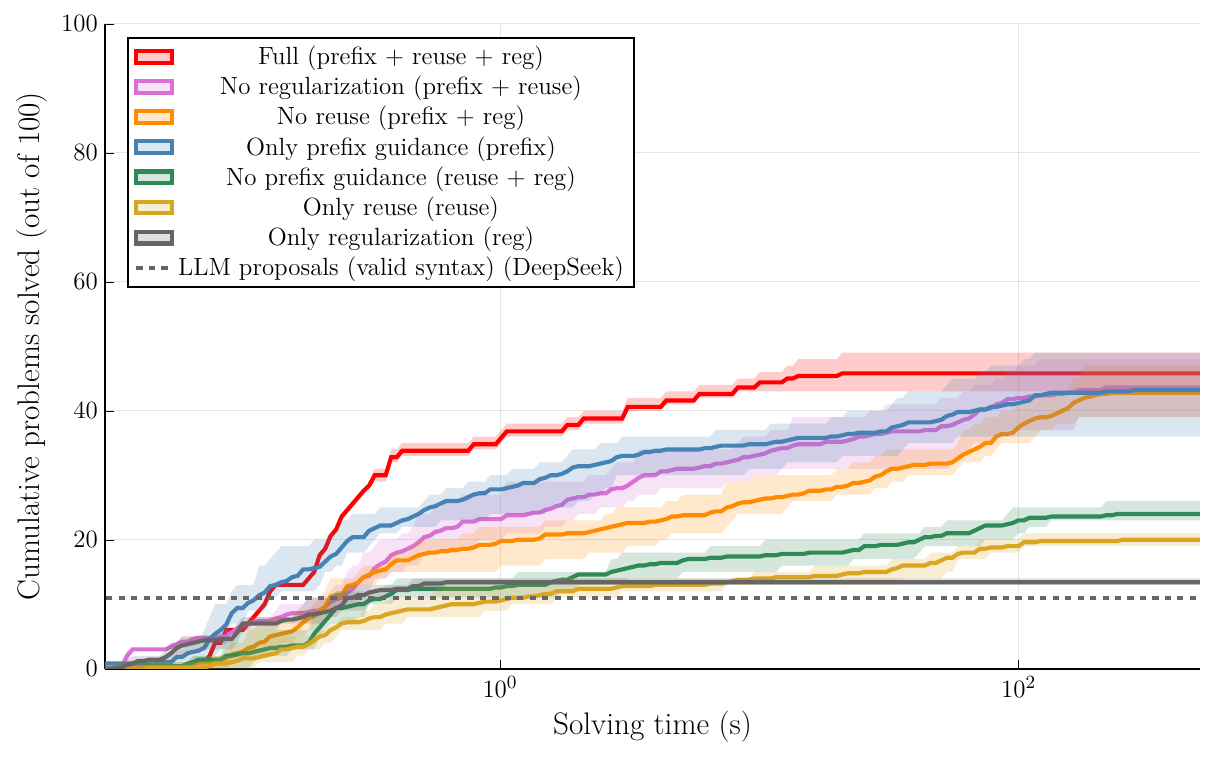}\\[4pt]
    \includegraphics[width=0.49\textwidth]{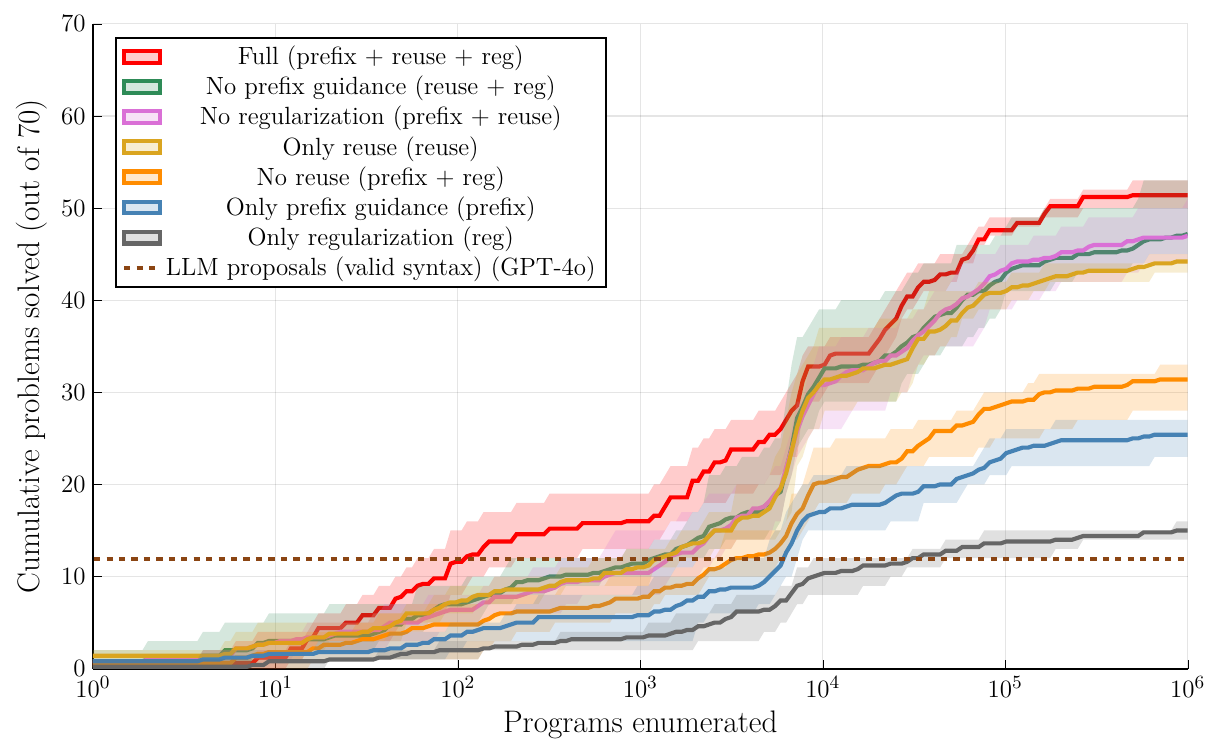}\hfill
    \includegraphics[width=0.49\textwidth]{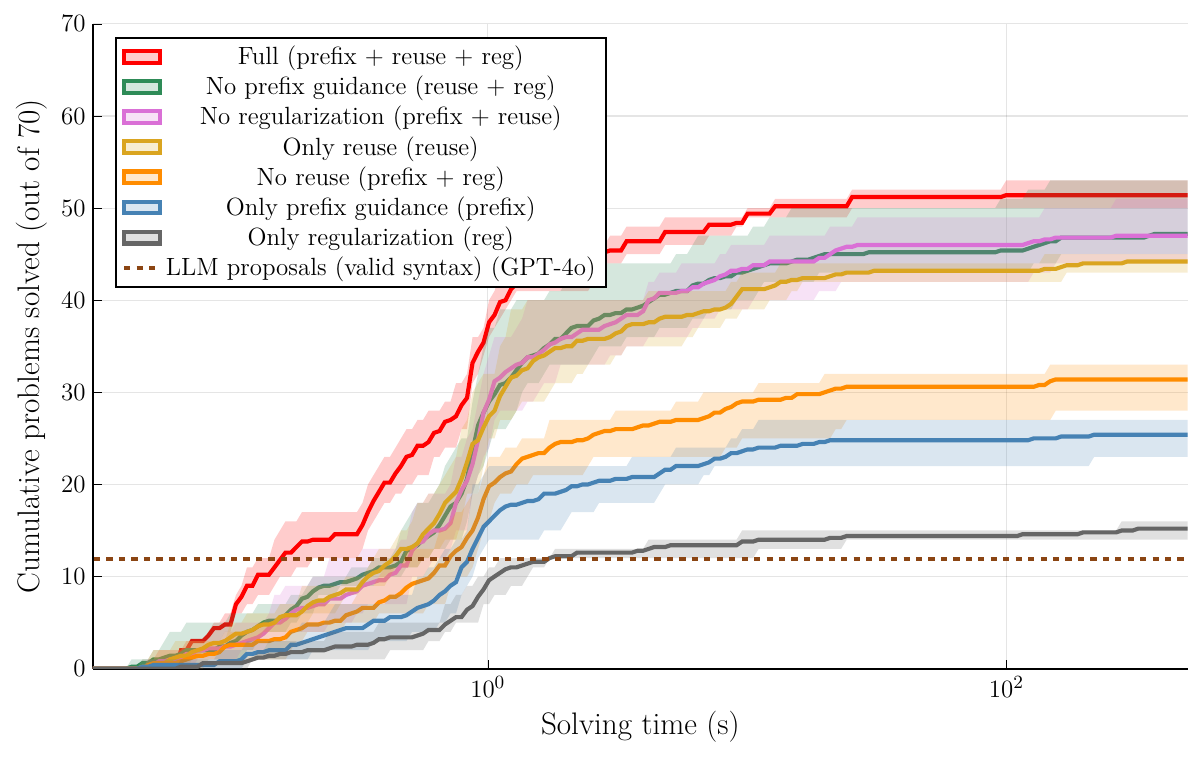}
    \caption{\textbf{RQ2: SLIA signal ablation (genetic backend) under weaker DeepSeek (top, 100 tasks) and stronger GPT-4o (bottom, 70 tasks) proposals, by programs enumerated (left) and wall-clock time (right).} Every signal contributes but none accounts for the full heuristic, and which signal is load-bearing inverts with proposal quality: under the weaker DeepSeek proposals prefix alignment dominates, since removing it costs the most and it alone recovers almost the full heuristic, whereas under the stronger GPT-4o proposals sub-program reuse takes over that role instead. This is why \ModelName{} carries all three: they cover the full range of proposal support rather than any one succeeding everywhere.}
    \label{fig:rq2-ablation}
\end{figure*}

Figure~\ref{fig:rq2-ablation} gives the full ablation discussed under RQ2, with one row per proposal model.
Note that the reuse-only variant is still not equivalent to the static heuristic: it also mines the recurring fragments into the grammar and scores those macro-rules, which a rule-frequency prior does not do.

\subsection{RQ1: ARC and ARGA}
\begin{figure*}[tbp]
    \centering
    \includegraphics[width=0.49\textwidth]{figures/plots/rq3_arga_cumulative_enumerations.pdf}\hfill
    \includegraphics[width=0.49\textwidth]{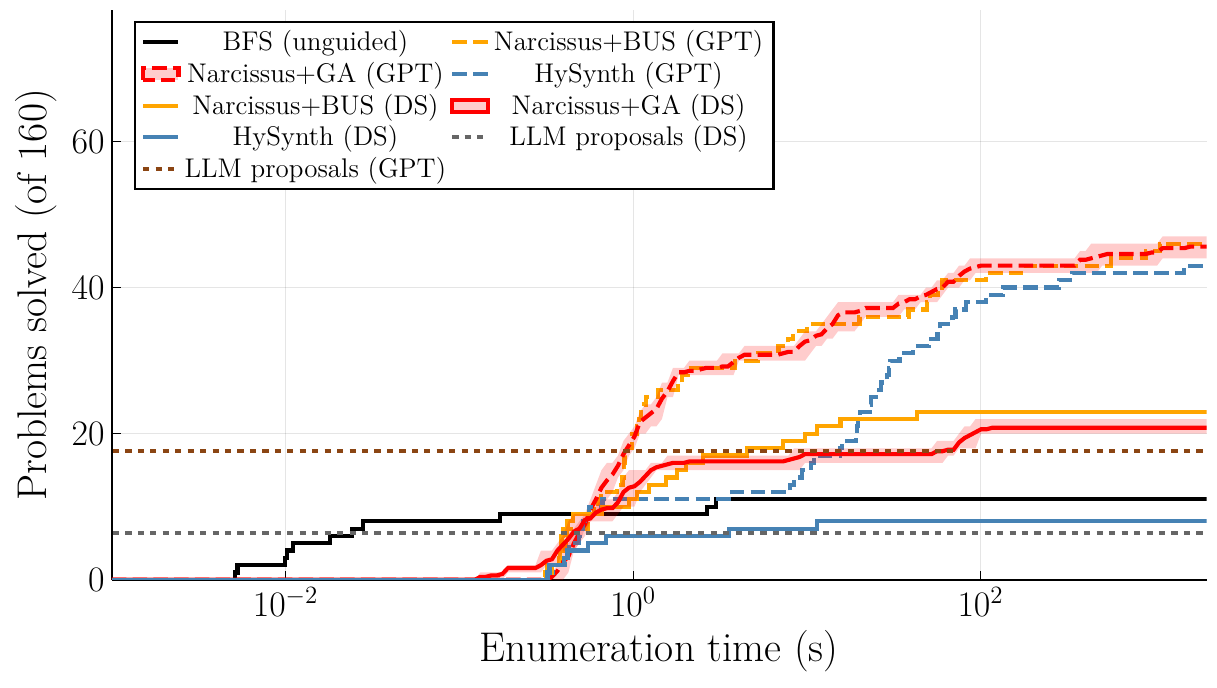}
    \caption{\textbf{RQ1, ARGA ($160$ tasks), GPT-4o vs.\ DeepSeek proposals: programs enumerated (left) and wall-clock time (right).} The left panel is the time-axis counterpart of Figure~\ref{fig:rq1}(b). Under divide-and-conquer, \ModelName{} beats HySynth for both proposal models and stays well above the raw proposals (dotted).}
    \label{fig:rq3-arga}
\end{figure*}

\begin{figure*}[tbp]
    \centering
    \includegraphics[width=0.49\textwidth]{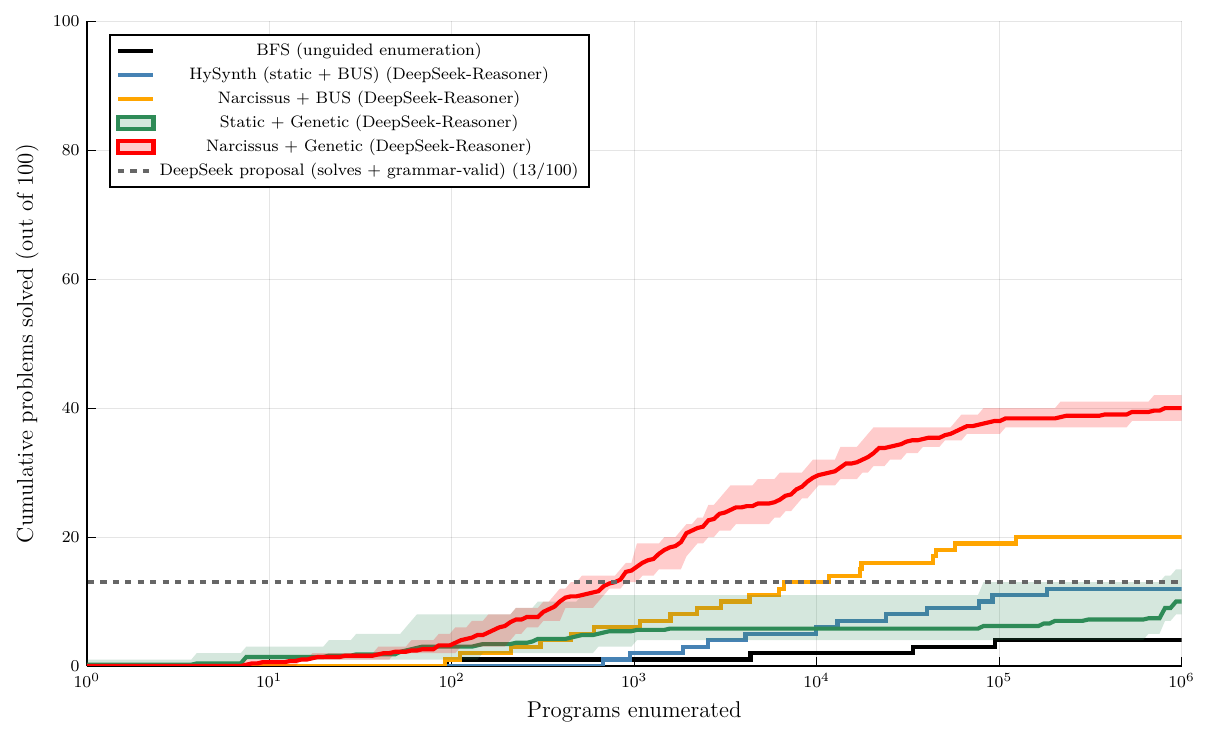}\hfill
    \includegraphics[width=0.49\textwidth]{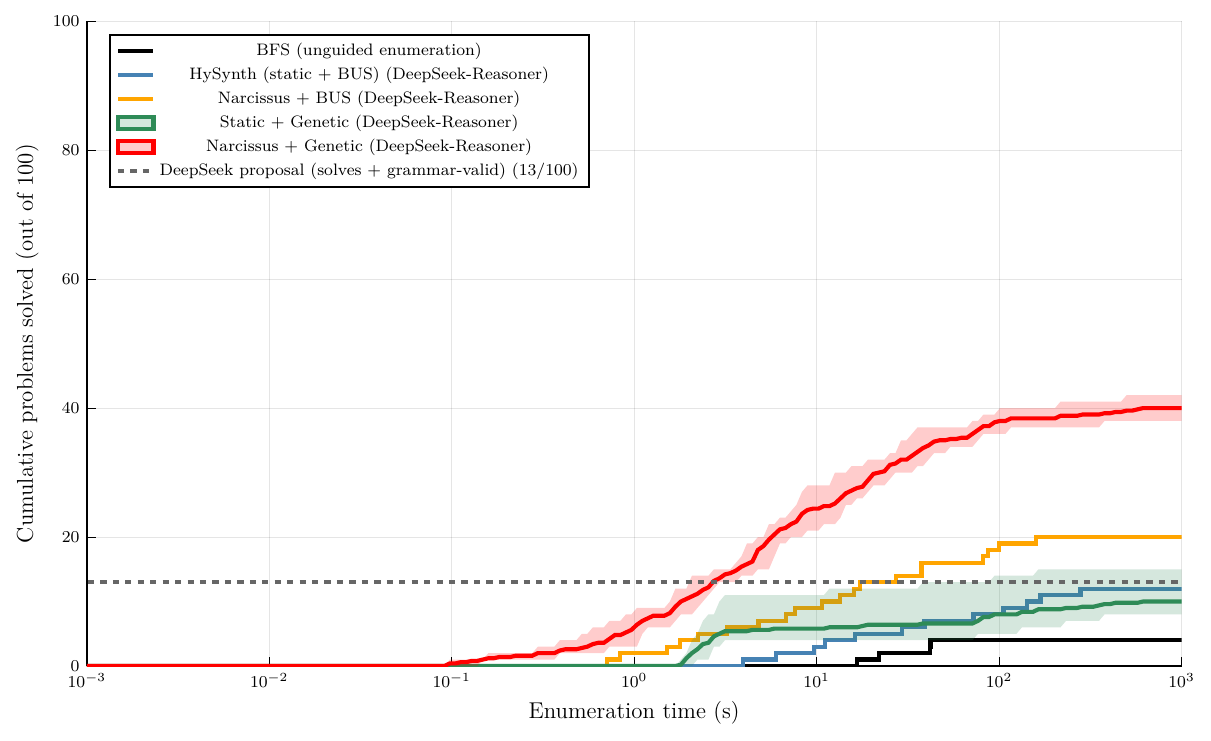}
    \caption{\textbf{RQ1, full ARC with the universal Hodel grammar (DeepSeek-Reasoner proposals): programs enumerated (left) and wall-clock time (right).} This is the weakest-support setting in the paper. \ModelName{} rises well above the dotted line marking the proposals that are both correct and grammar-valid, i.e.\ guided search solves ARC tasks for which not a single grammar-valid proposal is correct: the proposals contribute fragments and partial structure, and the search assembles and corrects them.}
    \label{fig:rq3-arc}
\end{figure*}

Figure~\ref{fig:rq3-arga} covers ARGA and is the wall-clock counterpart of Figure~\ref{fig:rq1}(b); Figure~\ref{fig:rq3-arc} covers full ARC over the Hodel grammar.

\section{AUC Tables}
\label{app:auc}

\subsection{How to Read the AUC Columns}
\label{app:auc-columns}

A cumulative solve curve plots the number of tasks solved (y) against a budget (x): either programs enumerated, from $1$ to $10^{6}$, or solving time, from $0.03$s to the $300$s timeout.
Summarizing such a curve by its endpoint alone discards exactly what we care about, namely \emph{how early} the tasks are solved, so we also report the area under the curve.
Because the budget spans orders of magnitude, the area is integrated in $\log_{10}x$ space, which weights each decade of budget equally instead of letting the last decade dominate, and then normalized by both the number of decades spanned ($6$ for the enumeration axis, $4$ for the time axis) and the benchmark size.
The generated tables report two columns per method:

\begin{description}
    \setlength{\itemsep}{2pt}
    \item[Final solved] The number of tasks solved once the whole budget is exhausted, i.e.\ the right-hand endpoint of the curve. Fractional values are means over five seeds.
    \item[AUC (\% of universe)] The normalized area under the curve, i.e.\ the \emph{average share of the benchmark solved} over the log-budget. $100\%$ means every task is solved instantly, and a steeper rise scores higher. This is the only quantity comparable across domains, since it is normalized for both benchmark size and budget range, and it is the column reported as ``AUC'' throughout the main text and in Table~\ref{tab:rq5-auc}.
\end{description}

\subsection{Full AUC Tables}

The following tables give both budget axes for every domain, reporting final solved count and normalized AUC as described above.
The enumeration axis for the SLIA LLM comparison is omitted here, since it is reproduced as Table~\ref{tab:rq5-auc} in the main text.

\begin{table}[h]
\centering
\footnotesize
\caption{RQ1, SLIA (DeepSeek) : Enumeration axis AUC (programs enumerated, 1 to $10^6$), out of 100 problems.}
\label{tab:auc-rq1-enum}
\resizebox{\linewidth}{!}{%
\begin{tabular}{lrr}
\toprule
Method & Final solved & AUC (\% of universe) \\
\midrule
Narcissus + Genetic (DeepSeek) & 41.8 & 22.1\% \\
HySynth (static + BUS) (DeepSeek) & 38.0 & 18.2\% \\
Static + Genetic (DeepSeek) & 28.4 & 17.8\% \\
Narcissus + BUS (DeepSeek) & 40.0 & 16.5\% \\
BFS (enhanced grammar) & 28.0 & 16.1\% \\
BFS (plain grammar) & 22.0 & 13.3\% \\
\bottomrule
\end{tabular}}
\end{table}

\begin{table}[h]
\centering
\footnotesize
\caption{RQ1, SLIA (DeepSeek) : Time axis AUC (solving time, 0.03s to 300s (timeout)), out of 100 problems.}
\label{tab:auc-rq1-time}
\resizebox{\linewidth}{!}{%
\begin{tabular}{lrr}
\toprule
Method & Final solved & AUC (\% of universe) \\
\midrule
Narcissus + Genetic (DeepSeek) & 41.8 & 31.2\% \\
Narcissus + BUS (DeepSeek) & 40.0 & 24.2\% \\
HySynth (static + BUS) (DeepSeek) & 38.0 & 20.7\% \\
Static + Genetic (DeepSeek) & 28.4 & 19.0\% \\
BFS (enhanced grammar) & 28.0 & 18.7\% \\
BFS (plain grammar) & 22.0 & 15.1\% \\
\bottomrule
\end{tabular}}
\end{table}

\begin{table}[h]
\centering
\footnotesize
\caption{RQ1, SLIA: LLM comparison (GPT-4o vs. DeepSeek) : Time axis AUC (solving time, 0.03s to 300s (timeout)), out of 70 problems.}
\label{tab:auc-sliallmcomparison-time}
\resizebox{\linewidth}{!}{%
\begin{tabular}{lrr}
\toprule
Method & Final solved & AUC (\% of universe) \\
\midrule
Narcissus + Genetic (GPT-4o) & 51.4 & 49.6\% \\
Narcissus + BUS (GPT-4o) & 49.0 & 47.3\% \\
HySynth (static + BUS) (GPT-4o) & 48.0 & 36.9\% \\
Narcissus + Genetic (DeepSeek) & 32.6 & 29.4\% \\
Static + Genetic (GPT-4o) & 32.2 & 25.0\% \\
Narcissus + BUS (DeepSeek) & 30.0 & 24.8\% \\
HySynth (static + BUS) (DeepSeek) & 28.0 & 16.3\% \\
Static + Genetic (DeepSeek) & 13.8 & 11.5\% \\
BFS (enhanced grammar) & 9.0 & 8.4\% \\
\bottomrule
\end{tabular}}
\end{table}

\begin{table}[h]
\centering
\footnotesize
\caption{RQ1, PBE\_BV\_Track\_2018 (bitvector), divide-and-conquer : Enumeration axis AUC (programs enumerated, 1 to $10^6$), out of 587 problems.}
\label{tab:auc-bvdc-enum}
\resizebox{\linewidth}{!}{%
\begin{tabular}{lrr}
\toprule
Method & Final solved & AUC (\% of universe) \\
\midrule
Narcissus + BUS & 350.0 & 33.2\% \\
Narcissus + Genetic & 349.4 & 28.8\% \\
BFS (unguided enumeration) & 300.0 & 27.5\% \\
HySynth (static + BUS) & 102.0 & 8.8\% \\
Static + Genetic & 56.6 & 4.8\% \\
\bottomrule
\end{tabular}}
\end{table}

\begin{table}[h]
\centering
\footnotesize
\caption{RQ1, PBE\_BV\_Track\_2018 (bitvector), divide-and-conquer : Time axis AUC (solving time, 0.03s to 300s (timeout)), out of 587 problems.}
\label{tab:auc-bvdc-time}
\resizebox{\linewidth}{!}{%
\begin{tabular}{lrr}
\toprule
Method & Final solved & AUC (\% of universe) \\
\midrule
Narcissus + BUS & 350.0 & 60.7\% \\
BFS (unguided enumeration) & 300.0 & 51.6\% \\
Narcissus + Genetic & 349.4 & 42.8\% \\
HySynth (static + BUS) & 102.0 & 16.0\% \\
Static + Genetic & 56.6 & 7.6\% \\
\bottomrule
\end{tabular}}
\end{table}

\begin{table}[h]
\centering
\footnotesize
\caption{RQ1, PBE\_BV\_Track\_2018 (bitvector), regular : Enumeration axis AUC (programs enumerated, 1 to $10^6$), out of 587 problems.}
\label{tab:auc-bv-enum}
\resizebox{\linewidth}{!}{%
\begin{tabular}{lrr}
\toprule
Method & Final solved & AUC (\% of universe) \\
\midrule
Narcissus + Genetic & 32.8 & 2.8\% \\
Narcissus + BUS & 26.0 & 2.6\% \\
BFS (unguided enumeration) & 27.0 & 2.5\% \\
Static + Genetic & 24.0 & 1.9\% \\
HySynth (static + BUS) & 18.0 & 1.8\% \\
\bottomrule
\end{tabular}}
\end{table}

\begin{table}[h]
\centering
\footnotesize
\caption{RQ1, PBE\_BV\_Track\_2018 (bitvector), regular : Time axis AUC (solving time, 0.03s to 300s (timeout)), out of 587 problems.}
\label{tab:auc-bv-time}
\resizebox{\linewidth}{!}{%
\begin{tabular}{lrr}
\toprule
Method & Final solved & AUC (\% of universe) \\
\midrule
BFS (unguided enumeration) & 27.0 & 3.8\% \\
Narcissus + Genetic & 32.6 & 3.4\% \\
Narcissus + BUS & 26.0 & 3.0\% \\
Static + Genetic & 24.0 & 2.3\% \\
HySynth (static + BUS) & 18.0 & 2.0\% \\
\bottomrule
\end{tabular}}
\end{table}

\begin{table}[h]
\centering
\footnotesize
\caption{RQ1, DeepCoder: LLM comparison (DeepSeek vs. Haiku) : Enumeration axis AUC (programs enumerated, 1 to $10^6$), out of 100 problems.}
\label{tab:auc-deepcoder-enum}
\resizebox{\linewidth}{!}{%
\begin{tabular}{lrr}
\toprule
Method & Final solved & AUC (\% of universe) \\
\midrule
Narcissus + Genetic (DeepSeek) & 32.6 & 19.8\% \\
Narcissus + Genetic (Haiku) & 30.6 & 19.7\% \\
Narcissus + BUS (DeepSeek) & 28.0 & 17.5\% \\
Narcissus + BUS (Haiku) & 26.0 & 16.5\% \\
HySynth (static + BUS) (Haiku) & 22.0 & 14.8\% \\
HySynth (static + BUS) (DeepSeek) & 20.0 & 13.6\% \\
Static + Genetic (DeepSeek) & 15.6 & 9.7\% \\
Static + Genetic (Haiku) & 13.8 & 9.2\% \\
BFS (unguided enumeration) & 10.0 & 6.2\% \\
\bottomrule
\end{tabular}}
\end{table}

\begin{table}[h]
\centering
\footnotesize
\caption{RQ1, DeepCoder: LLM comparison (DeepSeek vs. Haiku) : Time axis AUC (solving time, 0.03s to 300s (timeout)), out of 100 problems.}
\label{tab:auc-deepcoder-time}
\resizebox{\linewidth}{!}{%
\begin{tabular}{lrr}
\toprule
Method & Final solved & AUC (\% of universe) \\
\midrule
Narcissus + Genetic (DeepSeek) & 32.6 & 22.1\% \\
Narcissus + Genetic (Haiku) & 30.6 & 20.9\% \\
Narcissus + BUS (DeepSeek) & 28.0 & 18.9\% \\
Narcissus + BUS (Haiku) & 26.0 & 17.4\% \\
HySynth (static + BUS) (Haiku) & 22.0 & 15.6\% \\
HySynth (static + BUS) (DeepSeek) & 20.0 & 14.1\% \\
Static + Genetic (DeepSeek) & 15.6 & 10.9\% \\
BFS (unguided enumeration) & 10.0 & 10.0\% \\
Static + Genetic (Haiku) & 13.8 & 9.7\% \\
\bottomrule
\end{tabular}}
\end{table}

\end{document}